\documentclass{article}

\usepackage{microtype}
\usepackage{graphicx}
\usepackage{booktabs} %
\usepackage{wrapfig}
\usepackage{caption}
\usepackage{subcaption}
\usepackage{array}
\usepackage{multirow}
\usepackage{enumitem}
\usepackage{float}
\usepackage{placeins}

\usepackage{hyperref}

\newcommand{\name}{MM}
\newcommand{\namefull}{Marginal Matching}

\usepackage[accepted]{icml2025}
\makeatletter
\renewcommand{\ICML@appearing}{\textit{Preprint.}}
\makeatother
\hypersetup{pdfsubject={}}

\usepackage{amsmath}
\usepackage{amssymb}
\usepackage{mathtools}
\usepackage{amsthm}

\usepackage[capitalize,noabbrev]{cleveref}

\theoremstyle{plain}
\newtheorem{theorem}{Theorem}[section]

\newtheorem{lemma}[theorem]{Lemma}
\newtheorem{corollary}[theorem]{Corollary}
\theoremstyle{definition}
\newtheorem{definition}[theorem]{Definition}

\theoremstyle{remark}

\newtheorem{example}[theorem]{Example}

\icmltitlerunning{Correcting Flows with Marginal Matching}

\begin{document}

\twocolumn[
\icmltitle{Correcting Flows with Marginal Matching}

\icmlsetsymbol{equal}{*}

\begin{icmlauthorlist}
\icmlauthor{Meital Bojan}{tech}
\icmlauthor{Nadav Bojan Sellam}{tech}
\icmlauthor{Ev Zisselman}{tech}
\icmlauthor{Aviv Tamar}{tech}
\end{icmlauthorlist}

\icmlaffiliation{tech}{Technion -- Israel Institute of Technology, Haifa, Israel}

\icmlcorrespondingauthor{Meital Bojan}{meital.bojan@campus.technion.ac.il}

\icmlkeywords{Machine Learning, ICML}

\vskip 0.3in
]

\printAffiliationsAndNotice{} %

\begin{abstract}

Flow matching models, ODE-based generative models, generate samples by gradually morphing a simple source distribution into a target distribution. 
However, these models still face challenges in optimally balancing computational efficiency with generation fidelity. 
While predictor-corrector schemes have been proven to help address this issue in diffusion models effectively, their application to flow matching models remains under-explored.
In this work we present a novel training-based predictor-corrector framework for flow matching, which exploits the fact that the ODE can be easily reversed. Our key idea, termed \namefull, is to correct samples at a particular time step in the forward ODE towards a matching marginal distribution of the backward ODE.
This scheme proves to reduce a bound on the distance between the data distribution and the learned distribution, motivating two different implementations for the correction function. 
We show that our proposed method improves sample quality on CIFAR-10 and ImageNet-64, with minimal overhead in computation time.

\end{abstract}

\section{Introduction}

Flow-based generative models have emerged as a promising alternative to diffusion models, offering faster sampling while maintaining high generation quality \citep{lipman2023flow,liu2023flow,tong2024improving}. While these models require fewer sampling steps than diffusion probabilistic models (DPMs), they still face challenges in balancing computational efficiency with generation quality. This trade-off is crucial, as applications like image synthesis and audio generation require real-time performance, while protein design and robotics demand efficient resource utilization for complex tasks \citep{yan2024perflow,liu2023instaflow,le2024voicebox,irwin2024efficient, hu2023rfpolicy}.

In DPMs, predictor-corrector schemes have proven effective at improving sample quality. These schemes combine prediction steps, which estimate the next time step's sample, with correction steps that refine samples to better match the current time-step distribution. Previous work include both training-free methods \citep{song2021scorebased, zhao2023unipc,zhao2024dc}, which do not require additional model training, and training-based techniques \citep{lezama2022discrete}, which involve training a model specifically tailored to the task. However, applying these schemes to flow matching models remained relatively unexplored until recently. \citet{song2024leveraging} demonstrated the viability of implementing a predictor-corrector scheme for flow matching models without requiring additional training.

Traditional predictor-corrector frameworks aim to refine a sample at time step $t$ of the generation towards the marginal distribution of the learned diffusion process at time $t$.
For flow matching, the marginal distribution at time $t$ is the push-forward of the initial Gaussian distribution by the ODE up to time $t$.
However, due to limitations of the neural network training, a \textit{learned} flow does not exactly map Gaussian noise to perfect images, even with perfect integration.
Here, we introduce a novel training-based predictor-corrector approach specifically tailored for flow matching models that leverages known marginal distributions to provably yield the exact data distribution. 
Our method exploits time reversibility, which has been explored in diffusion models \citep{wallace2023edict,mokady2023null}, but is arguably less involved in flow matching models as it only requires solving the ODE in reverse time \citep{liu2023flow}.

Consider a reverse ODE, starting from the data samples and flowing back in time according to the learned flow model. Our main insight is that starting the generation from samples on this trajectory will generate perfect images. Building on this, we enhance generation quality through a predictor-corrector scheme that alternates between ODE solver steps (predictor) and learned correction steps, reducing the discrepancy between generated samples and the time-reversal ODE solution. 
We provide theoretical guarantees that such a correction function minimizes a bound on the Wasserstein distance \citep{kantorovich1942translocation} between the target approximation and the true target distribution, as a function of the error reduction of the corrections at different time steps. We hypothesize that for practical flow models, the forward and backward ODEs will be relatively aligned, making the correction easy to learn and apply with minimal overhead.

We examine two practical implementations for the correction function.  
The first implementation minimizes a theoretical bound, but has access to only subset of the data during training. The second has access to the full dataset, but makes no theoretical guarantee.  We examine these correction models empirically and perform an extensive ablation study for our design choices.  
Our results show significant improvement with minimal additional steps.

In summary, our \textbf{main contributions} are as follows:
\vspace{-2ex}
\begin{itemize}
    \setlength{\itemsep}{0pt}
    \item A novel corrector model for predictor-corrector sampling of flow matching models based on time reversal.
    \item A theoretical analysis of this general framework, and two practical implementations.
    \item We empirically show that our proposed correction models improve results on CIFAR-10 and ImageNet-64 datasets, achieving lower FID scores with fewer sampling steps. 
\end{itemize}

\section{Background}

\subsection{Flow Models}\label{section:flow_models}
Given a source distribution $\pi_0$ and a target distribution $\pi_1$, both over $\mathbb{R}^d$, flow matching \citep{lipman2023flow,liu2023flow,tong2024improving} models their mapping.  This is done with a smooth time-dependent vector field $u:[0,1]\times\mathbb{R}^d \rightarrow \mathbb{R}^d$ which defines an ODE $d\phi_t(x) = u_t(\phi_t(x)) dt$, where $x \in \mathbb{R}^d$, $t \in [0,1]$ and $\phi_t$ is the flow map with the initial condition $\phi_0(x)=x$.
A time-varying density function $p:[0,1]\times\mathbb{R}^d \rightarrow \mathbb{R}_{>0}$ that satisfies the continuity equation:
\vspace{-1.5ex}
\begin{align}
\small
    \frac{\partial p_t}{\partial t} + div(p_t(x)u_t(x)) = 0,
\label{eq:continuity_equation}
\end{align}
with the initial density $p_0$ over $\mathbb{R}^d$ is called the \textit{marginal probability path} generated by the vector field $u$, and $u$ is said to be the \textit{probability flow ODE} for the density function $p$.

The flow matching (FM) objective regresses $v_\theta:[0,1]\times\mathbb{R}^d \rightarrow \mathbb{R}^d$, a time-dependent vector field parametrized as a neural network with weights $\theta$, to a target vector field $u_t$ via the following mean squared error (MSE) loss function:
\vspace{-1.5ex}
\begin{align}
    \mathcal{L}_{FM}(\theta)=\mathbb{E}_{t\sim[0,1],x\sim p_t(x)}\|v_\theta(t,x)-u_t(x)\|^2.
\label{eq:flow_matching}
\end{align}
\vspace{-3ex}

Numerous mappings exist between probability distributions $\pi_0$ and $\pi_1$.  The distributions $p_t(x)$ and $u_t(x)$ are not unique and typically impossible to compute directly.
To address this, \citet{lipman2023flow} introduced the conditional flow matching (CFM) loss, which provides equivalent gradients to Eq.~\ref{eq:flow_matching} with respect to $\theta$.
\vspace{-1.5ex}
\begin{align}
  \small   
    \mathcal{L}_{CFM}(\theta)=\mathbb{E}_{t\sim[0,1],z\sim q(z),x\sim p_t(x|z)}\|v_\theta(t,x)-u_t(x|z)\|^2,   \label{exp:conditional_flow_matching}
\end{align}
\vspace{-2ex}

where $z$ is some conditional variable sampled from a distribution $q(z)$.
 The full derivation is available in \citet{lipman2023flow,tong2024improving}. Many efficient parameterizations exist for $u_t(x|z)$ and $p_t(x|z)$, in this work we focus on the parametrization described in \citet{tong2024improving}:
\begin{align*}
    z =(x_0,x_1) \quad ; \quad 
    u_t(x|z) = x_1-x_0 \quad ; \\
    p_t(x|z)=\mathcal{N}(x|t\cdot x_1 + (1-t) \cdot x_0, \sigma^2I),
\end{align*}
where $z\sim q(z)$ and $t\in[0,1]$ is the interpolation coefficient. The conditional density $p_t(x|z)$ specifies one of the possible $p_t$ distributions and is easy to sample from and learn. 
The vector field $u_t$ and its corresponding marginal distribution $p_t$ can be obtained in terms of the conditional ones:
    $p_t(x) = \int p_t(x|z)q(z) dz$, 
    $u_t(x) = \int \frac{p_t(x|z)u_t(x|z)}{p_t(x)}q(z) dz$.
 We examine two recently studied flow models  by \citet{tong2024improving}, one with independent coupling (I-CFM) $q(z)=\pi_0(x_0)\times \pi_1(x_1)$ and another with optimal transport coupling (OT-CFM) $q(z)=OT(x_0,x_1)$. The optimal transport (OT) problem maps $x_0$ to $x_1$ such that a displacement cost, (typically the squared Euclidean distance), is minimized, for more details see Appendix.~\ref{appendix:optimal_transport}.  The OT coupling is calculated on batches during training and results in straighter trajectories, \citep{pooladian2023multisample,tong2024improving}.

\subsection{Score Models}\label{section:score_models}
In score-based generative models \citep{song2021scorebased, song2019generative} the forward and reverse processes are modeled using standard diffusion and reverse-time stochastic differential equations (SDEs). The forward process is described by $dx = f(x,t)dt + g(t)dw$, while the reverse-time SDE is given by $dx = \left(f(x,t) - g(t)^2 \nabla \log \pi_1\right)dt + g(t)dw$. In these equations, $w$ represent standard Wiener processes, $f$ is the drift coefficient, $g$ is the diffusion coefficient, and $t$ ranges from $0$ to $T_s$. The reverse SDE is used for generating samples, evolving from $t=T_s$ to $t=0$, for example by using a numerical SDE solver.

The score model estimates the score function $\nabla_x \log {\pi_1(x)}$, where $\pi_1(x)$ is the target distribution defined above. The training process typically involves denoising score matching (DSM):
\vspace{-1.5ex}
\begin{align}
  \small   
L_{DSM}(\theta) = \mathbb{E}_{\pi_1(x)p_\sigma(\Tilde{x}|x)}{\|s_\theta(\Tilde{x}) - \nabla_{\Tilde{x}} \log{ p_\sigma(\Tilde{x}| x)\|^2_2} }
\label{eq:denoised_score_matching}
\end{align}
Here, $s_\theta$ denotes a neural network parameterized by $\theta$, and $p_\sigma(\Tilde{x}|x)=\mathcal{N}(x,\sigma^2I)$ represents the probability of a noisy sample from the target distribution. 
\citet{vincent2011connection} demonstrated that the optimal score network $s_{\theta^*}(x)$, which minimizes Eq.~\ref{eq:denoised_score_matching}, satisfies $s_{\theta^*}(x) = \nabla_x \log {p_\sigma(x)}$ almost surely.  This result shows that the network learns to estimate the score function of $p_\sigma(x)$, which is crucial for effectively denoising it.
However, the approximation $s_{\theta^*}(x) = \nabla_x \log {p_\sigma(x)} \approx \nabla_x \log {\pi_1(x)}$ holds true only when the noise level is sufficiently small, such that $p_\sigma(x) \approx \pi_1(x)$.

\subsection{Predictor-Corrector Methods}

\begin{algorithm}[H]
\footnotesize
    \centering
    \caption{General Predictor-Corrector Inference}\label{alg:predictor_corrector}
    \begin{algorithmic}[1]
    \REQUIRE {number of prediction disretization steps $N$, number of correction steps $M$}
    \STATE $x_0 \sim \mathcal{N}(0,I)$
    \FOR{$n = 0$ to $N-1$}
    \STATE $x_{t_{n+1}} = \text{Predictor}(x_{t_{n}})$
    \FOR{$m = 1$ to $M$}
        \STATE $x_{t_{n+1}} = \text{Corrector}(x_{t_{n+1}})$
    \ENDFOR
    \ENDFOR
    \STATE \textbf{return} $x_{t_N}$
    \end{algorithmic}
\end{algorithm}

\vspace{-2ex}

Predictor-corrector approaches, as shown in \citep{zhao2023unipc, lezama2022discrete, song2021scorebased}, are common in DPMs, where a corrector is intermittently applied during inference to improve performance (Algorithm~\ref{alg:predictor_corrector}). 
Previous work often employs the same pre-trained model for both prediction and correction, differing only in whether the model predicts the next step or refines the current one.
For instance, \citet{song2021scorebased} use a score model with an ODE solver for prediction and Langevin Dynamics for correction. Meanwhile, \citet{song2024leveraging} store flow model outputs in a buffer for use during correction. Unlike previous approaches, we propose using a dedicated correction model and investigate its design.

\section{Method}

\subsection{Motivation and Problem Formulation}

Given a frozen pre-trained flow model $v_\theta$, our goal is to improve the sampling of $v_\theta$. In this paper, we focus on the correction phase of the predictor-corrector framework and propose training an auxiliary correction model to enhance the approximation of the marginal distributions along the inference trajectory. 
We term our predictor-corrector method \namefull\ (\name). 

\begin{figure*}%
\centering
 \begin{subfigure}{0.15\textwidth}
     \includegraphics[width=\textwidth]{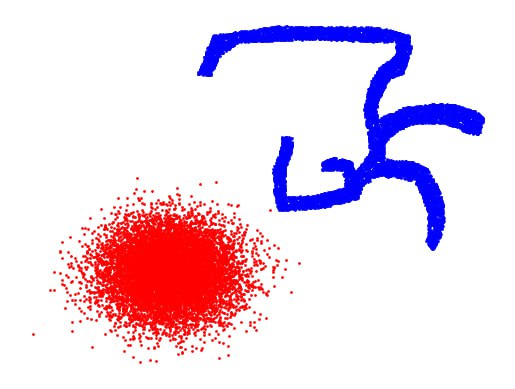}
     \caption{$\pi_0$ and $\pi_1$}
     \label{fig:source_target}
 \end{subfigure}
 \hspace{7ex}
 \begin{subfigure}{0.15\textwidth}
     \includegraphics[width=\textwidth]{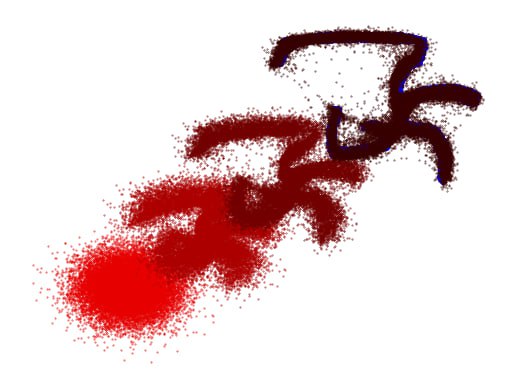}
     \caption{Forward}
     \label{fig:forward_marginals}
 \end{subfigure}
\hspace{7ex}
 \begin{subfigure}{0.15\textwidth}
     \includegraphics[width=\textwidth]{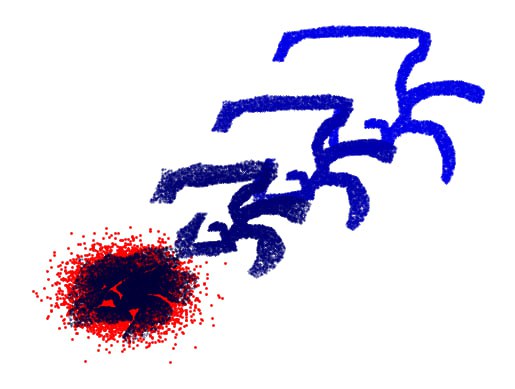}
     \caption{Backward}
     \label{fig:backward_marginals}
 \end{subfigure}
    \caption{\textbf{Illustration of Marginals} 
    (a) Gaussian source distribution $\pi_0$ (red), and target distribution $\pi_1$ (blue).
    (b) Discrete forward marginals $\{p^f_{t_n}\}_{n=0}^N$.
    (c) Discrete backward marginals $\{p^b_{t_n}\}_{n=0}^N$.
    Note that $p_{t_n}^b \neq p_{t_n}^f$ for every $n$, specifically $p_0^b$ (=$\hat{\pi}_0$, black) differs from $p_0^f$ (=$\pi_0$, red).}
    \label{fig:toy_exp}
\vspace{-3ex}
\end{figure*}

Langevin Dynamics \citep{song2021scorebased} is a common correction method, but its implementation becomes challenging with flow models that do not inherently represent a score function. A potential solution is to train an auxiliary score model to capture every intermediate marginal distribution. During prediction, the flow model advances to the next time-step, while the correction phase leverages the score model to more accurately approximate the marginal distributions. 
The key challenge lies in defining these marginals.
An initial approach would be to use the marginal distributions the flow model was originally trained on— an interpolation between the target distribution and Gaussian noise. 
In the subsequent discussion, we will explore an alternative choice for these marginal distributions.

The flow model $v_\theta$ is trained to map a source distribution $\pi_0$ to a target distribution $\pi_1$\footnote{
In practice, $\pi_1$ is not available during training; only samples from it are. In our discussion here we do not make a distinction between the ideal data distribution and the empirical data distribution that the finite data prescribes. Empirically, we did not observe issues of over-fitting to the training data.}, but in practice it maps to an approximation of $\pi_1$. Let $\hat{\pi}_1$ denote the distribution of samples generated by $v_\theta$, which does not exactly match $\pi_1$.

If the vector field represented by $v_\theta$ is invertible, a well defined ``reverse flow'' exists from the target distribution according to the reverse of the vector field, $-v_\theta$. The key observation is that starting a forward flow using $v_\theta$ from any point on the reverse flow will converge to the target distribution $\pi_1$. 
That is, we have found a trajectory of densities that works perfectly with our imperfect model $v_\theta$. We assume that the flow is invertible, implying cycle consistency between forward and backward vector field mappings \citep{zhu2017unpaired}.
A similar assumption is used in \citet{liu2023flow} in experiments on image translation. While we cannot verify that this assumption holds in practice, we empirically validate our method on popular datasets.

We next formally define the forward and backward marginals, representing the time-evolving probability distributions generated by the vector field.

\begin{definition} \label{definition:forward_backward_marginals}
    \textbf{Forward and Backward Marginals}: Let $u$ be a continuous globally Lipschitz time-dependent function $u:[0,1]\times\mathbb{R}^d \rightarrow \mathbb{R}^d$.
    Let $u^{-1}$ be the reverse vector field defined as: $u^{-1}(t,x):=-u(t,x)$. The discretization of the time-interval $[0,1]$ into $N$ steps is $t_n=n/N $.
    \vspace{-2ex}
    \begin{itemize}
        \item \textit{Forward Marginals} 
        are the probability path generated by $\frac{\partial p^f_t}{\partial t} + div(p^f_t(x)u_t(x)) = 0$ from time $0$ to $1$, with initial distribution $p_0^f=\pi_0$. The final distribution is $p_1^f=\hat{\pi}_1$. Their $N$ steps discretization is defined as $\{p^f_{t_n}\}_{n=0}^N$.
        \item  \textit{Backward Marginals} 
         are the probability path generated by $\frac{\partial p^b_t}{\partial t} + div(p^b_t(x)u^{-1}_t(x)) = 0$ from time $1$ to $0$, with initial distribution $p_1^b=\pi_1$.  The final distribution is $p_0^b=\hat{\pi}_0$. Their $N$ steps discretization is defined as $\{p^b_{t_n}\}_{n=0}^N$\footnote{To ensure clarity, we use consistent time indexing for all marginals: subscript $0$ represents the initial point of the forward marginals and the final point of the backward marginals.}.
    \end{itemize}
\end{definition}

We refer to $p_0^b$ as the approximate source distribution to distinguish it from $\pi_0$, the source distribution, and $p_1^f$ as the approximate target distribution to distinguish it from $\pi_1$, the target distribution. 
Fig.~\ref{fig:toy_exp} illustrates the forward and backward marginals of a flow model sampled with $10$-steps Euler integration. It demonstrates the probability marginals trajectory and that $p_{t_n}^b \neq p_{t_n}^f$ for every $n$. Additionally, there is a substantial intersection between the forward and backward marginals, and their symmetric difference represents the outliers that do not reach the target distribution, for more details see Appendix.~\ref{appendix:forward_vs_backward}. Therefore, we predict that refining samples towards the marginals $v_\theta$ was trained on will be beneficial, though less so than backward marginals, as supported by the experiments.

Our next observation is that if the forward marginals trajectory reaches the target distribution, then it aligns with the backward trajectory for every time-step.
This is formally stated in Lemma.~\ref{lemma:forward_backward_equality}

\begin{lemma}\label{lemma:forward_backward_equality}
Let $u: [0, 1] \times \mathbb{R}^d \to \mathbb{R}^d$ be a continuous time-dependent vector field, satisfying:
$|u_t(x) - u_t(y)| \leq L |x - y|$, $\forall t \in [0, 1]$, $x, y \in \mathbb{R}^d$. Then,
$p^f_1=p^b_1=\pi_1$ if and only if $p^f_t=p^b_t$ $\forall t\in[0,1]$.
\end{lemma}

\begin{figure}%
\centering
    \includegraphics[width=0.45\textwidth]{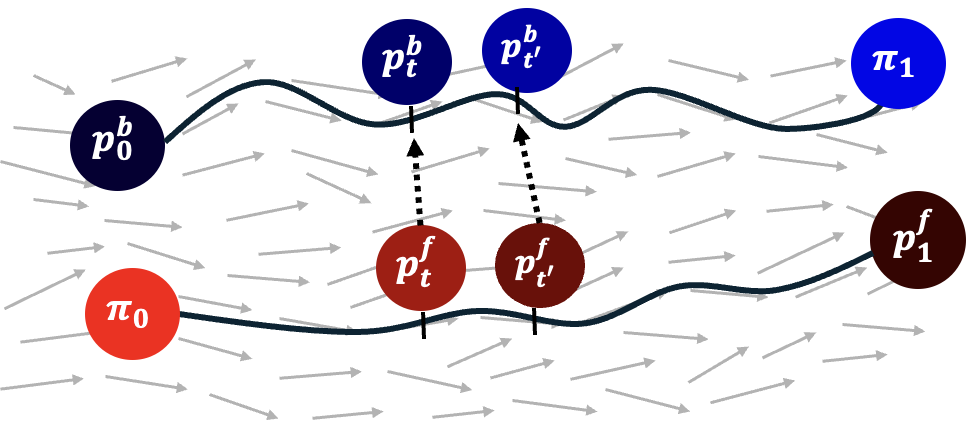}
  \caption{\textbf{Marginals in Vector Field} The black lines show two key trajectories in a vector field. The lower is the forward trajectory, starts at $\pi_0$ and ends at $p_1^f=\hat{\pi}_1$. The upper is the backward trajectory, begins at $p_0^b$ and ends at the data distribution $\pi_1$. Both are discretized at time-steps $t$ and $t'$. Our objective is to transition from forward to backward marginals trajectory.}
  \label{fig:trajectories_in_vector_field}
\vspace{-3ex}
\end{figure}

Based on this observation, we suggest refining the generated samples to better align with the backward marginals, ensuring that the final sample distribution more effectively matches the target distribution. Fig.\ref{fig:trajectories_in_vector_field} illustrates this idea.

Practically, the backward marginals can be obtained by running an ODE-solver in reverse (from time $1$ to time $0$) on the frozen pre-trained model $v_\theta$ starting at the data samples, (see Appendix.~\ref{appendix:backward_generation}). Then, a correction model can be trained to bridge between the forward and backward marginals.

\subsection{Correction Steps Prioritization}\label{section:steps_prioritization}

In the prediction-correction scheme, correction steps are typically performed after each prediction step. While these steps can enhance performance, they also increase computational cost. The following theorem indicates that not all correction steps are equally significant, suggesting that to balance performance and computation time, we should prioritize the correction steps that offer the greatest benefit.

\begin{theorem} \label{thm:informal_trajectory_divergence_reductions_bound_prob} (Informal)
Let $u$ be a well-behaved vector field with Lipshitz constant L, $p^f_t$ and $p^b_t$ be two time-dependent probability density functions satisfying the continuity equation Eq.~(\ref{eq:continuity_equation}) with respect to $u$. Suppose that the initial Wasserstein distance is $d_0 = W_2(p^b_0, p^f_0)$. At each time step $t_n$, a correction function $h$ is applied to $p^f_{t_n}$ and the flow function continues from $h(t_n, p^f_{t_n})$. Assuming $h$ reduces the $W_2$ distance to $p^b_{t_n}$ by $\epsilon_n$, the reduction on the bound of the final distance is: $W_2(p^b_1, p^f_1)\leq d_0 \exp\{L\} - \sum_{i=0}^{n-1} {\epsilon_i \exp\{L(1 - t_i)\}}$.
\end{theorem}

A formal theorem statement and proofs are available in Appendix.~\ref{appendix:proofs}, which also includes calculations for two specific error reduction functions—additive and multiplicative. The following examples demonstrate that in each case the total error reduction is spread differently across the time steps, illustrating an important idea -- we should prioritize the corrections applied in particular steps based on the error reduction model. We will revisit this idea in our experiments.

\label{thm_examples}
\begin{example} \textbf{Linear Reduction}
This corresponds to the case of linear error reduction in Thm.~\ref{thm:informal_trajectory_divergence_reductions_bound_prob}. For $N=2$, the bound is: $d_0\exp\{L\}-\epsilon_0\exp\{L\}-\epsilon_1\exp\{L\cdot 0.5\}$.  Assuming $\epsilon_0=\epsilon_1$, the reduction term at step $0$ has a greater impact on the final bound due to the larger exponential factor. This highlights the importance of prioritizing corrections at earlier steps in the sequence.
\end{example}

\begin{example} \textbf{Multiplicative Reduction}
This corresponds to the case of multiplicative error reduction in Thm.~\ref{thm:informal_trajectory_divergence_reductions_bound_prob}. For $N=2$, the bound is: $d_0\exp\{L\}(1-\epsilon_0)(1-\epsilon_1)$.  Here, the reduction terms $(1-\epsilon_0)$ and $(1-\epsilon_1)$ contribute equally to the final bound, irrespective of their position in the sequence. Thus, the optimal strategy is to prioritize improvements at steps with the largest $\epsilon_i$.
\end{example}

Thm.\ref{thm:informal_trajectory_divergence_reductions_bound_prob} states that the transport map $h$ can be utilized to reduce the distance between the forward $p^f_t$ and backward marginals $p^b_t$. In the following, slightly abusing notation, we assume that $h$ is the push-forward of $x\in \mathbb{R}^d$ with a mapping $h(x)$.
In practice, the specific error reduction is unknown and we can only require that the bound on the Wasserstein distance will decrease after $h$ is applied. 

Let $h_\psi$ be a neural network with weights $\psi$ that approximates the correction function $h$. 
We consider two models for implementing the correction model $h_\psi$.

\subsection{Correction Models Implementation}

\subsubsection{Score Model}

\vspace{-1ex}
\begin{algorithm}[H]
    \centering
    \caption{Score Model Training}\label{alg:score_model_training}
    \begin{algorithmic}[1]
    \REQUIRE {flow model $v_\theta$, score model $h_\psi$, $\sigma$, backward marginals $\{p^b_{t_n}(x)\}_{n=0}^N$}
    \REPEAT
    \STATE $\epsilon \sim \mathcal{N}(0,I)$
    \STATE $n\sim U(\{0,1,..,N\})$
    \STATE $x_{t_n} \sim p_{t_n}^b(x)$
    \STATE Take gradient descent step on
    \STATE $\nabla_\psi \| h_\psi(t_n, x_{t_n} + \sigma\cdot \epsilon) + \epsilon\|^2$
    \UNTIL {converged}
    \end{algorithmic}
\end{algorithm}
\vspace{-2.5ex}

Our first proposed approach is a time-dependent score model $h_\psi(t,x_t)=s_\psi(t,x_t)$, presented in Algorithm.~\ref{alg:score_model_training}, to approximate the backward marginals $\nabla_{x_t} \log{p^b_{t}}(x_t)$. 
We assume this score function is well-defined when evaluated on samples from the forward marginal distributions (i.e. $x_t\sim p^f_t$), which allows us to use it to move between these distributions (our experiments provide empirical support for this assumption).
Motivated by the bound of \citet{kwon2022score} (Thm. $2$) for score models generation, the following lemma
bounds the final Wasserstein distance after applying the score correction on a single time-step marginal:
\vspace{-1ex}
\begin{lemma}(Informal)\label{lemma:one_step_score}
   Applying $s_\psi$ from $p^f_{t_n}$ to approximate $p^b_{t_n}$ as described in Sec.~\ref{section:score_models}
   bounds the final distance $W_2(p^b_1,p^f_1)$ as follows:
   \begin{align*}
       W_2(p^b_1,p^f_1) \le C(s_\psi)\exp\{L(1-t_n)\},
   \end{align*} 
\end{lemma}
\vspace{-2ex}
where $C(s_\psi)$ depends on the score model's loss $L_{DSM}^\psi$ as defined in Sec.~\ref{section:score_models}. 
For small enough $L_{DSM}^\psi$, this bound improves upon using no correction.
For more details and a bound on the general case---applying score correction on multiple marginals, see 
Appendix.~\ref{appendix:score_wasserstein}.

In practice, we noticed that a single correction step is enough, and sampling with Langevin Dynamics is unnecessary. This correction step can be done with any gradient that points in the direction of the backward marginals.

\subsubsection{Robust Classifier}\label{section:classifier}

In the score model, we did not use data from the forward marginals during training. We next propose a method that explicitly transitions between samples from the forward and backward marginals.
 This approach utilizes a time-dependent classifier $c_\psi(t,x_t)$ that distinguishes between forward and backward marginals ($p_t^f$ and $p_t^b$), evaluating each sample independently. By learning to categorize samples as belonging to either the forward or backward marginal at each time step, the classifier's gradients $h_\psi(t,x_t)=\nabla_{x_t} \log {c_\psi(t,x_t)}$ provide corrective guidance, effectively steering samples toward the backward marginals. The gradients will be zero upon reaching the correct class. 

To enhance the classifier's gradient performance, we implement two key improvements: Adversarial Training (\textbf{AT}) and Gradient Alignment (\textbf{GA}).  These keep the classifier's gradients stable and meaningful, for more details and the complete algorithm for training see Appendix.~\ref{appendix:classifier_training}.

\textbf{Usage for Class Conditioning} 
Classifier guidance is used in diffusion models and more recently in flow models \citep{sun2024rectifid} in order to generate conditional images. 
While \name\ generates data samples, it can also extend to conditional data generation. 
For this purpose, our classifier can be conditioned $c_\psi(t,x_t,l)$ where $l$ is a class label, and trained to categorize different classes in the backward trajectory. For an illustration see Appendix.~\ref{appendix:classifier_guidance}.

\section{Related Work}

\textbf{Diffusion predictor-corrector methods} -- 
Predictor-corrector methods have gained significant attention for balancing computational efficiency and sampling accuracy in diffusion processes. First introduced by \citet{song2021scorebased}, this approach uses a score model for both prediction and correction, employing an SDE/ODE solver for prediction and Langevin dynamics to refine marginal distributions for correction.
UniPC \citep{zhao2023unipc} improved upon \citet{lu2022dpm} by leveraging a unified predictor-corrector framework that incorporates previous time-step predictions. Subsequently, \citet{zhao2024dc} introduced further refinements with a dynamic compensation strategy. These methods are all training-free, as they utilize the same model for both prediction and correction. For discrete data, \citet{lezama2022discrete} proposed a method that uses pre-trained base models in conjunction with a learned corrector to enhance synthesis performance. Similarly, we adopt an approach that trains a correction model based on a pre-trained base model. However, it is important to note that while these methods are designed for DPMs, our work focuses on flow models, which operate under a different framework.

\textbf{Acceleration methods for flow models} -- 
Acceleration methods for flow-based models have focused on improving ODE-solving efficiency. Concurrent with our work, \citet{song2024leveraging} introduced a predictor-corrector method based on a training-free flow solver that minimized function evaluations by caching prior results and leveraging Taylor expansions to approximate ODE solutions efficiently. Rectified flow \citet{liu2023flow}, demonstrated an acceleration approach through reflow operations to straighten probabilistic paths and further refined this process by training a distillation model over the final result. Moreover, \citet{shaulbespoke} proposed a training-based acceleration technique that adapts flow-based models into DPMs for faster computation. Despite these advancements, the exploration of ODE-based acceleration techniques specifically tailored for flow matching models remains underdeveloped compared to their DPMs counterparts due to their unique probabilistic path representation.
For extended related work see Appendix.~\ref{appendix:extened_related_work}.

\begin{table}%
\vspace{-2ex}
\small
\begin{minipage}[t]{0.46\textwidth}
\centering
\setlength{\tabcolsep}{4pt}
\begin{tabular}{ l l  c  c}
&&&\\
  \textit{Model}& Sampler &NFE  $\downarrow$ & FID $\downarrow$  \\ 
 \hline
 DDPM  & Adaptive& 274  & 7.48 \\
 Score Matching& Adaptive & 242  & 19.94\\
 OT-FM & Adaptive& 142  & 6.35\\
 OT-CFM  & RK4& 40  & 4.34\\
 & Adaptive& 133.94 & 3.58\\
 I-CFM & RK4 & 40 & 4.29\\
 & Adaptive& 146.42 & 3.66 \\
 \hline
\end{tabular}
\end{minipage}%
\hspace{0.3em}%
\begin{minipage}[t]{0.48\textwidth} 
\centering
\setlength{\tabcolsep}{1pt}  %
\begin{tabular*}{\linewidth}{@{\extracolsep{\fill}}l l l c c@{}}
&&&OT-CFM&I-CFM
 \\
  \textit{Ours} & NFE $\downarrow$ & C-NFE $\downarrow$ & FID $\downarrow$ & FID $\downarrow$ \\ 
 \hline
  Score & 41 &1 ~~\textit{($n=10$)} & 3.45 & 3.47\\
 & 43 &3 ~~\textit{($n=0,5,10$)} & 3.38 & 3.39\\
 & 51  &11 \textit{($n=0\dots10$)}  &\textbf{3.37} & \textbf{3.38}\\
 Classifier& 42 &1 ~~\textit{($n=8$)} & 3.57 & 3.77\\
 & 46 & 3 ~~\textit{($n=8,9,10$)}& 3.48 & 3.67\\
 & 50 & 5 ~~\textit{($n=6,7,8,9,10$)} & \textbf{3.47} & \textbf{3.62}\\
 & 62 & 11 \textit{($n=0\dots10$)} & 3.48& 3.63\\
 \hline
\end{tabular*}
\end{minipage}
\caption{CIFAR-10 performance. (Top) Leading baseline models. (Bottom) Our correction models (score and classifier) with OT-CFM and I-CFM base flow models. The base flow models are sampled with 10-step RK4 (40 NFE). Correction steps improve performance with minimal additional NFEs, and generally, increasing the number of applied corrections (C-NFE) enhances overall results. Additional comparisons are available in Appendix.~\ref{appendix:additional_rk4}.}   
\label{table:cifar10_correction_models}
\vspace{-5ex}
\end{table}

\section{Experiments}

In this section, we empirically analyze the performance of \name's image generation under varying marginals, comparing score correction model and classifier correction model. We then investigate the performance under interpolated and backward distributions, followed by an examination of the effects of sampling a score correction model using Langevin Dynamics versus a single-step approach.
The score model is trained with a constant noise $\sigma$. We conduct experiments on two datasets: unconditional CIFAR-10 (\citet{krizhevsky2009learning}) and ImageNet-64 (\citet{chrabaszcz2017downsampled,deng2009imagenet}).  To assess the quality of generated images, we employ Fréchet Inception Distance (FID) (\citet{heusel2017gans}) score as our evaluation metric.

A key factor in diffusion and flow models is inference compute time, measured by number of function evaluations (NFE). In all tables, NFE represents the total number of sampling steps, and C-NFE denotes the steps taken with our correction model (out of the NFE).  We note that a classifier correction step require two NFEs, due to input derivative calculations, while score correction requires only one NFE.

For the flow model we use the UNet architecture with the same hyper-parameters as \citet{tong2024improving}. For the score and classifier models, we use UNet with half the number of parameters used for the flow model. 
Additionally, these correction models undergo significantly fewer training iterations compared to the flow model.
Unless otherwise specified, the correction models were trained on trajectories consisting of $N=11$ marginals with OT-CFM \citep{tong2024improving} as the flow model, and the flow model was sampled with 10-step RK4, which is 40 NFEs.  For more implementation details, please refer to the Appendix.~\ref{appendix:implementation_details}. The code will be available upon publication.

\begin{wrapfigure}{r}{0.25\textwidth}
\vspace{-3ex}
\centering
  \includegraphics[width=0.25\textwidth]{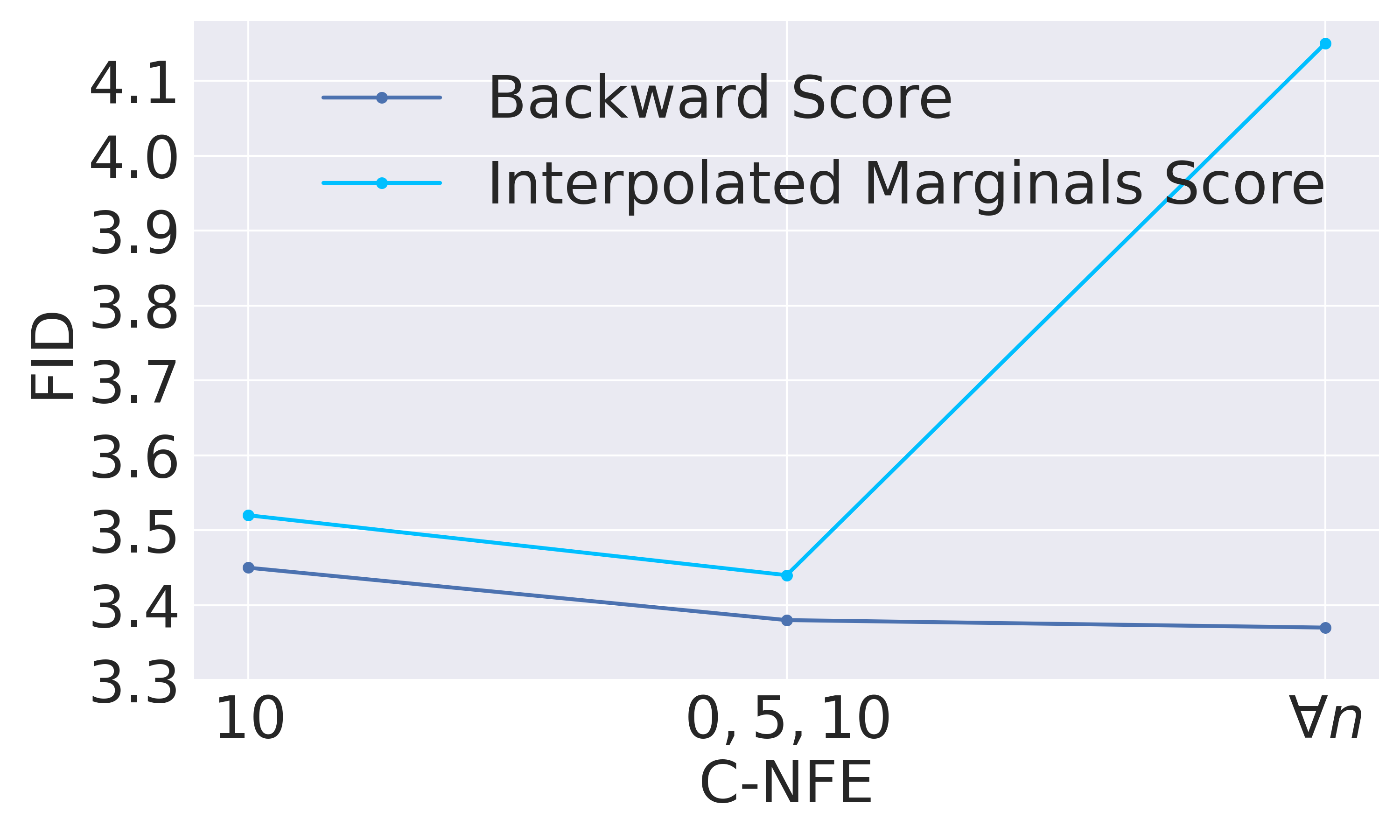}
\caption{CIFAR-10 sample quality comparison of score correction  trained on the interpolated ($\pi_1$ and $\pi_0$) vs. backward marginals $p^b_t$. The base flow is sampled with 10-step RK4.}
\label{fig:forawrd_backward_score}
\vspace{-4ex}
\hspace{-4ex}
\end{wrapfigure}

\subsection{CIFAR-10 Sample Quality}

\textbf{Score Correction vs. Classifier Correction Model} 
To showcase the improvement of \name\, we implement our score and classifier correction models with OT-CFM and I-CFM base flow models on CIFAR-10. Table~\ref{table:cifar10_correction_models} shows comparison of our results to \citep{ho2020denoising,song2021scorebased,lipman2023flow,tong2024improving}. Previous works employ an adaptive sampler (DOPRI5, \citet{dormand1980family}), while we use RK4 as the ODE-sampler to ease the integration of the correction steps. 
We compare with results reported in \citet{lipman2023flow}, implemented using the same UNet \citep{dhariwal2021diffusion}, and those in \citet{tong2024improving}. As \citet{tong2024improving} did not report RK4 sampling results, we conduct these experiments and report baseline results using the models' checkpoints as our foundational flow models. Following the results of Sec.~\ref{section:steps_prioritization}, we investigate applying the correction on various time steps (C-NFE $> 1$)\footnote{Multiple corrections helped the classifier. However, for simplicity we apply one correction step per marginal.}.
We examine several options: for the score model the correction steps were taken on $n=\{[10], [0,5,10], [0\dots10]\}$ and for the classifier on $n=\{[8], [8,9,10], [6,7,8,9,10], [0\dots10]\}$.

\begin{table}%
\small
\begin{minipage}[t]{0.48\textwidth}
\centering
\setlength{\tabcolsep}{4pt}
\begin{tabular}{ l c c c }
 \textit{Model}&NFE $\downarrow$ & Sample & FID $\downarrow$ \\
 \hline
 DDPM  & 264& Adaptive  & 17.36 \\
 Score Matching & 441& Adaptive  & 19.74 \\
 OT-FM  & 138& Adaptive  & 14.45 \\
 OT-CFM & 20& RK4  & 17.99\\
 & 24& RK4  & 16.60\\
  & 28& RK4  & 15.71\\
 \hline
\end{tabular}
\end{minipage}%
\hspace{0.5em}%
\begin{minipage}[t]{0.48\textwidth}
\centering
\setlength{\tabcolsep}{4pt}
\begin{tabular}{ l l l c }
 \textit{Ours} & NFE $\downarrow$ & C-NFE $\downarrow$ & FID \\
 \hline
 \multirow{3}{*}{Score} &21&1 \textit{($n=5$)}& 15.91\\
 &22&2 \textit{($n=0,5$)}& \textbf{15.53}\\
 &26&6 \textit{($n=0\dots5$)}& 15.57 \\
 \hline
\end{tabular}
\end{minipage}
\caption{ImageNet-64 results for top generative models (Top) vs. score correction with OT-CFM base flow (Bottom), which is sampled with 5-step RK4 (20 NFE).}
\label{table:imagenet_correction_models}
\vspace{-5ex}
\end{table}

We significantly outperform the baselines with  minimal additional NFEs. The score model slightly outperforms the classifier in FID score, despite not being exposed to forward marginals during training.
It also demonstrates consistent performance regardless of whether it is applied to OT-CFM or I-CFM, indicating a robust denoising ability from forward to backward marginals. Whereas, the robust classifier is sensitive  to the model choice, showing superior performance with OT-CFM. We did not observe any benefit from applying both the score and classifier correction models together.

\begin{figure*}
\centering
 \includegraphics[width=0.8\textwidth]{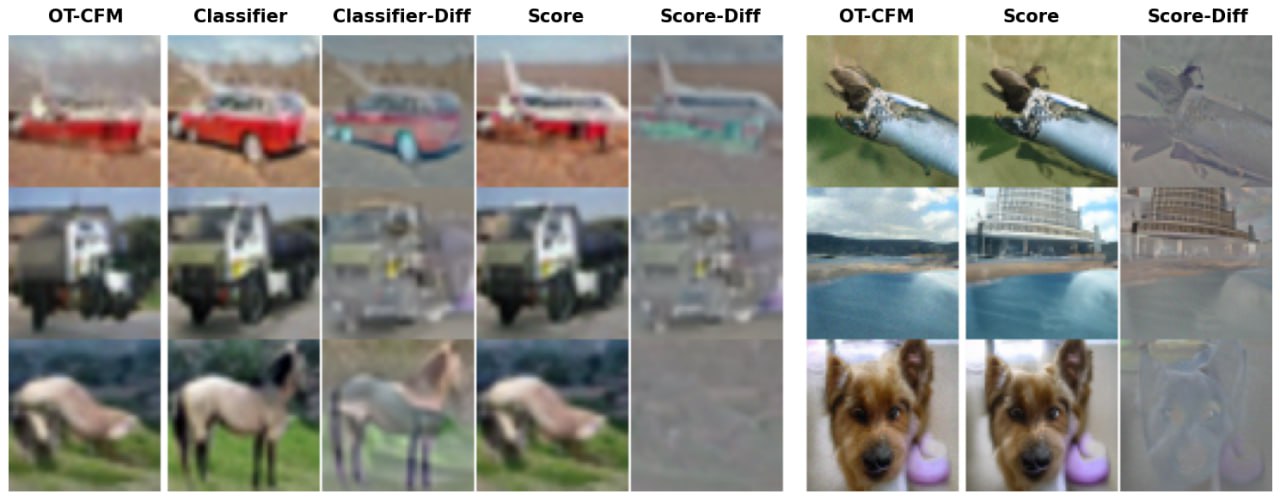}
     \caption{Correction models with OT-CFM as the base flow model on CIFAR-10 (Left) and ImageNet-64 (Right). The "-Diff" shows the difference between corrected and uncorrected images. Corrected images show improved sharpness and better alignment with the dataset.}
\label{fig:image_correction_comparison}
\vspace{-2ex}
\end{figure*}

Fig.~\ref{fig:image_correction_comparison} (Left) presents a qualitative comparison of the correction performed by each algorithm, (additional images are available in Appendix~\ref{appendix:imagenet_images}).  
The correction models improve clarity and sometimes make significant changes to the image. The score and classifier corrections show varying levels of effectiveness depending on the specific image. In some cases, one method outperforms the other, while in other instances, they surprisingly produce similar results.

\textbf{Interpolated Marginals vs. Backward Marginals} 
Fig.~\ref{fig:forawrd_backward_score} compares two correction approaches: refining toward interpolated versus backward marginals using the score model. 
Both corrections improve results, but correcting toward backward marginals performs better as they guide the process toward the true target distribution $\pi_1$. Using all correction steps for marginal interpolation steered the flow off course, resulting in degraded performance.

\textbf{Langevin Dynamics vs. One-Step Sampling} 

\begin{wrapfigure}{r}{0.25\textwidth}
\vspace{-1ex}
\centering
  \includegraphics[width=0.25\textwidth]{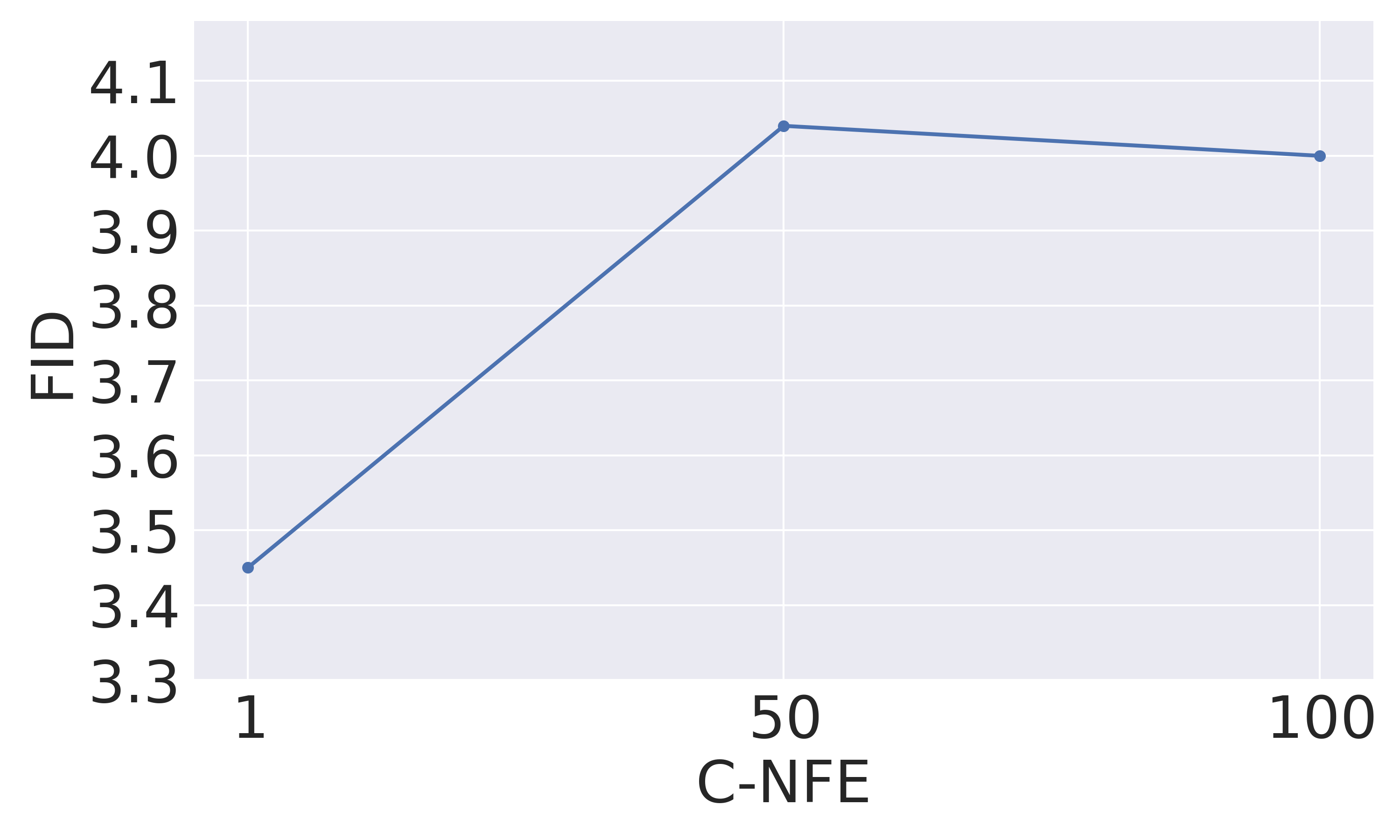}
\caption{CIFAR-10 sample quality of score correction sampled with $50$ and $100$ step Langevin Dynamics versus a single correction step at the $10^{\text{th}}$ time-step marginal. The base flow is sampled with 10-step RK4.}
\label{fig:score_langevin_vs_one}
\vspace{-5ex}
\hspace{-3ex}
\end{wrapfigure}

Fig.~\ref{fig:score_langevin_vs_one} compares two approaches for the correction score model trained on backward marginals: using multi-step Langevin Dynamics (with 50 and 100 steps) versus applying a single correction step at the $10^{\text{th}}$ time-step marginal (as used in Table.~\ref{table:cifar10_correction_models}). The multi-step Langevin Dynamics approach exhibited limitations - small step sizes resulted in minimal image changes, while larger steps introduced noise. The single correction step proved both computationally more efficient and produced better results.

\subsection{ImageNet-64 Sample Quality}\label{imagenet_result}
We perform an experiment on ImageNet-64 to examine \name\ in a higher-dimensional settings.  Given a pre-trained OT-CFM flow model trained on ImageNet-64, we train a score correction model to improve its sample quality, as the score model outperform the classifier on CIFAR, we chose to perform these resource-intensive experiments with it.  The score model was trained on trajectories consisting of $N=6$ marginals. Table.~\ref{table:imagenet_correction_models} presents the improvement in FID of a correction model trained on the backward marginals trajectory 
compared to \citep{ho2020denoising,song2021scorebased,lipman2023flow,tong2024improving}. Similarly to the CIFAR-10 results, we observe that \name\ improves FID with minimal addition of NFEs. The correction is qualitatively demonstrated in Fig.~\ref{fig:image_correction_comparison} (Right), (for more illustrations see Appendix.~\ref{appendix:imagenet_images}).  The correction model removes the residual noise in the model's approximation and changes the images structure to better align with the ImageNet dataset photos.

\subsection{Ablation Study}\label{ablation_study}

\subsubsection{Marginals Ablation}\label{marginal_ablation}

\begin{figure*}[t]%
\centering
 \includegraphics[width=0.27\textwidth]{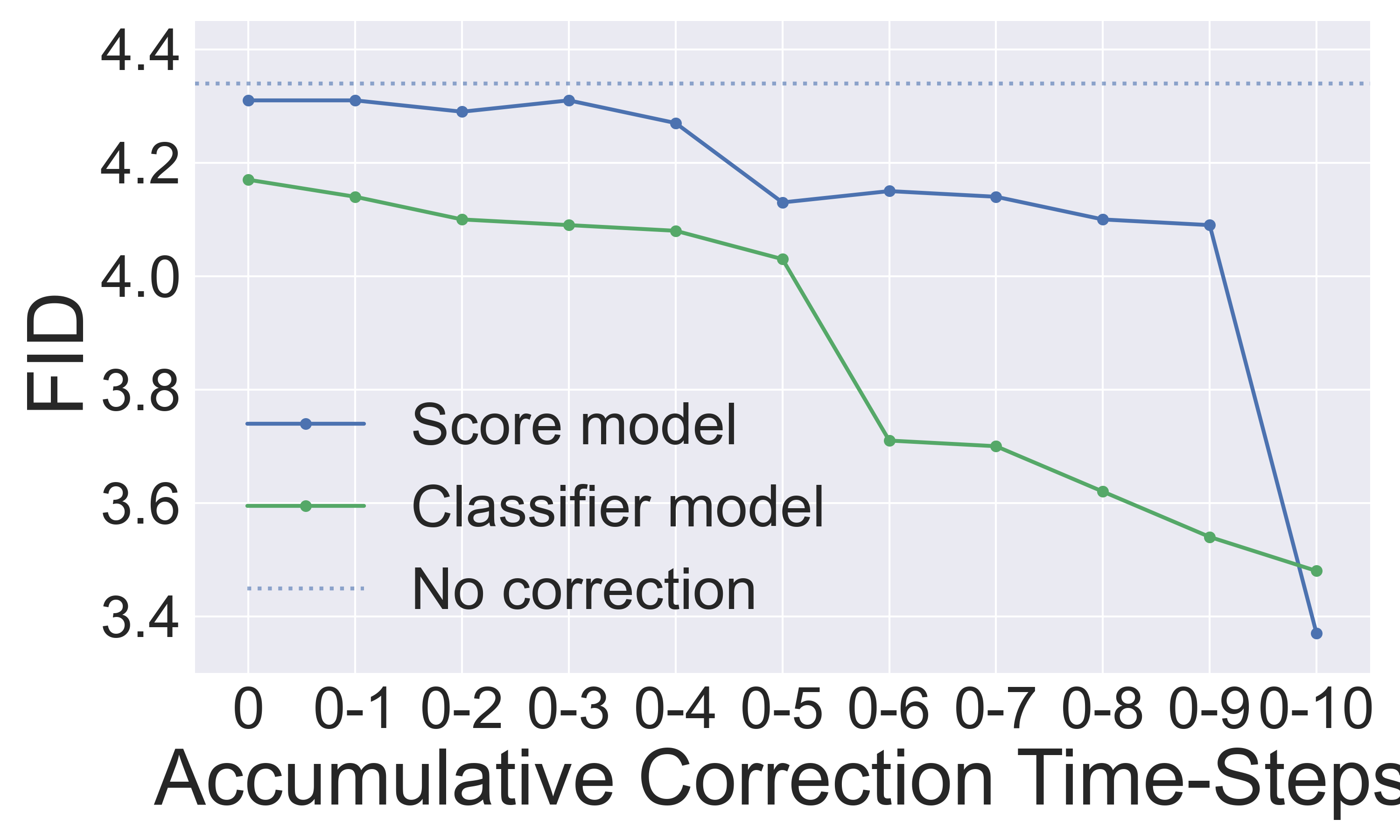}
  \includegraphics[width=0.27\textwidth]{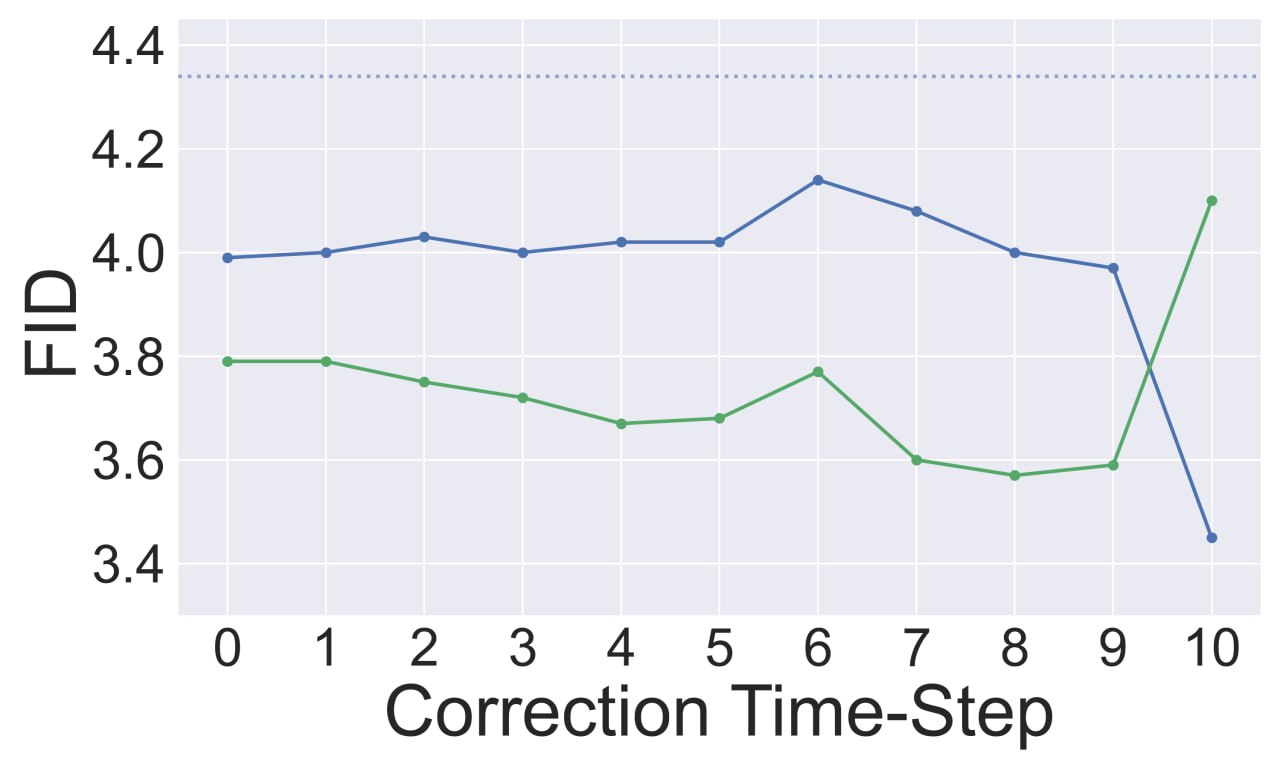}
\raisebox{-2.7pt}{\includegraphics[width=0.425\textwidth]{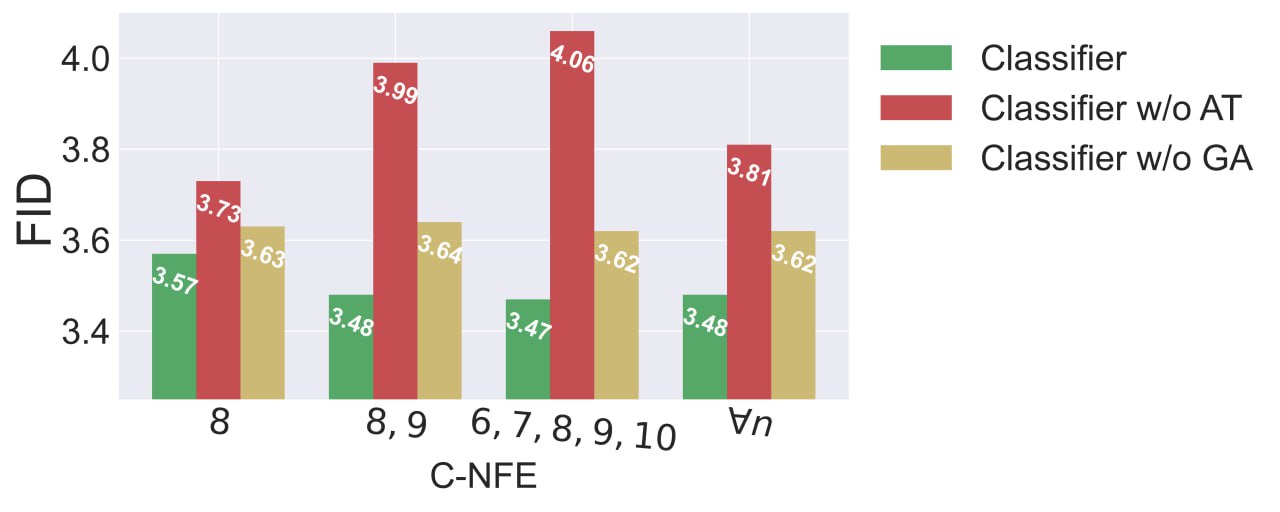}}

\caption{CIFAR-10 sample quality. The base flow model is OT-CFM sampled with 10-step RK4. \textbf{Left:} FID vs
accumulative correction time-steps. \textbf{Middle:} FID vs single correction time-step.  \textbf{Right:} Ablation study on classifier loss components sampled across different marginals.}
\label{fig:ablations}
\vspace{-3ex}
\end{figure*}

In this ablation we study the effect of applying correction on different time-steps. 
Fig.~\ref{fig:ablations} (Left) demonstrates the cumulative effect of correcting all time-step marginals (from 0 to 10). The score model demonstrates significant improvements primarily during the middle and final steps, with minimal contributions from early stages. In contrast, the classifier exhibits a more uniform pattern of enhancement.  
Fig.~\ref{fig:ablations} (Middle) presents the improvement in FID when applying a single correction step on different time-step marginals. 
The classifier model's correction consistently improves more than the score model at every time-step, except the final one.  
As predicted by the theoretical analysis in Sec.~\ref{section:steps_prioritization}, we observe that different correction algorithms yield varying levels of improvement when applied at different time steps.

Notably, early-stage corrections mainly affect low-frequency components, shaping overall image structure. In contrast, later-stage corrections primarily impact high-frequency details, addressing noise-like elements \citep{kim2024leveraging}.
The striking impact of executing the last time step with the score model suggests residual noise in the model's predictions (see Appendix.~\ref{appendix:imagenet_images} for illustrations).

\subsubsection{Classifier Loss Ablation}\label{classifier_loss_ablation}
 To assess the significance of each improvement to the classifier, namely Adversarial Training (AT) and Gradient Alignment (GA), we conduct an ablation study by removing these components. The results are presented in Fig.~\ref{fig:ablations} (Right). Without AT, the gradients become unstable, as seen from the inconsistent improvement with additional steps. Without GA, the gradients are small, and adding steps does not improve performance.

 For additional results, 
 see Appendix.~\ref{appendix:additional_experiments} and Appendix.~\ref{appendix:approximations}.

\section{Conclusion and Limitations}

This paper introduces a novel predictor-corrector framework for flow matching models, featuring a dedicated correction model. 
Our main insight is that given an imperfect flow model, a trajectory that exactly reaches the data distribution can be computed by reversing its vector field.
Based on this insight, we propose \namefull, a simple algorithm that steers the inference-time trajectory to better align with the path that accurately reaches the data.  
Unlike conventional predictor-corrector, which typically refine samples by aligning with the interpolated marginal distribution at each step, our approach directs samples toward a marginal distribution that steers toward the true data distribution. 
We demonstrate superior performance on CIFAR-10 and ImageNet-64, along with an extensive ablation study.

Ideally, all generative models  could potentially improve with a correction phase where refinement targets a marginal distribution that leads to the data distribution. A major limitation of our work is the assumption on the reversibility of the vector field, 
constraining our method from generalizing to non-reversible models. This limitation may be relaxed for approximately reversible models such as diffusion models \citep{wallace2023edict}. In general, extending our approach to non-reversible models is an interesting future direction.

Another aspect is the practical implementation of our theory. In Thm.~\ref{thm:informal_trajectory_divergence_reductions_bound_prob} we assumed knowledge of error reductions.  In practice, we do not have access to the model's error reduction (see Sec.\ref{thm_examples}).  
However, we can bound that reduction in the case of score correction (Lemma.~\ref{lemma:one_step_score}) and propose a bound on the final Wasserstein distance.
Future work could explore theoretical bounds also for our classifier-based correction model.

Finally, our method requires training an additional model after a flow model has been pre-trained, adding extra compute and parameters. It is worth noting that in all our experiments, this auxiliary model is smaller than the original flow model, and training it takes significantly less time. An alternative approach could involve training the flow and correction models together, either as a single model with both flow and correction outputs or as two models trained simultaneously.

\section*{Impact Statement}

This paper presents work whose goal is to advance the field of 
Machine Learning. There are many potential societal consequences 
of our work, none which we feel must be specifically highlighted here.

\bibliography{example_paper}
\bibliographystyle{icml2025}

\newpage
\appendix
\onecolumn

\section{Optimal Transport} \label{appendix:optimal_transport}

Optimal Transport (OT) is a mathematical framework that addresses the problem of efficiently moving probability mass from one distribution to another while minimizing a certain cost.  The central idea is to find the optimal way to transform one probability distribution into another, considering the "cost" of moving each unit of mass.

In the context of flow matching, before learning the flow, an optimal transport map is computed between batches of samples from $\pi_0$ and $\pi_1$. This OT step uses the 2-Wasserstein distance as its metric, with a cost function defined as the Euclidean distance $\|x-y\|$ between points. The flow is then learned based on these optimally transported samples, resulting in straighter trajectories \citep{pooladian2023multisample,tong2024improving}.

\section{Theoretical Proofs}
\label{appendix:proofs}

\begin{proof}[Proof of Lemma \ref{lemma:forward_backward_equality}] \label{proof:forward_backward_equality}
If $p_t^f=p_t^b$ for every $t$ then $p_1^f=p_1^b=\pi_1$ is trivial.

Assume $p^f_1=p^b_1=\pi_1$. We need to show that $p^f_t=p^b_t$ for every $t\in[0,1]$.

The flow $\phi_t$ is a diffeomorphism for each $t\in[0,1]$ due to the properties of the vector field $u$ (continuous and globally Lipschitz).

For any $t\in[0,1]$, we can write:
\begin{align*}
    p^f_t = [\phi_t]_\#p^f_0 = [\phi_t \circ \phi_1^{-1}]_\#p^f_1, \\
    p^b_t = [\phi_t]_\#p^b_0 = [\phi_t \circ \phi_1^{-1}]_\#p^b_1,
\end{align*}

where $\phi_1^{-1}$ is the inverse map of $\phi_1$ flowing from $t=1$ to $t=0$. Since the following equality holds: $p^f_1 = p^b_1 = \pi_1$, we can replace $p^f_1$ and $p^b_1$ with $\pi_1$ in the above equations:
$p^f_t = [\phi_t \circ \phi_1^{-1}]_\#\pi_1$
$p^b_t = [\phi_t \circ \phi_1^{-1}]_\#\pi_1$.

\end{proof}

\begin{lemma}\label{lemma:bound_distance_t}
Let $\phi_t$ be the flow map of the vector field $u$ that induces the push-forward $q^1_t := [\phi_t]_\#q^1_0$, $q^2_t := [\phi_t]_\#q^2_0$ 
for any two probability density distributions $q^1_0$ and $q^2_0$ over $\mathbb{R}^d$, where $t \in [0, 1]$. The 2-Wasserstein distance between $q^1_t$ and $q^2_t$ satisfies:
\[
W_2(q^1_t, q^2_t) \leq W_2(q^1_0, q^2_0) \exp\{Lt\}.
\]
\end{lemma}

\begin{corollary}
Assume $t'>t$: 
$W_2(q^1_{t'}, q^2_{t'}) \leq W_2(q^1_t, q^2_t) \exp\{L(t'-t)\}$.
\end{corollary}

\begin{proof}
Let $\tilde{o}_0$\footnote{Literature commonly uses the symbol $\pi$ to denote an optimal transport coupling. This paper, however, already uses $\pi$ to represent source and target distributions. To avoid confusion, $\tilde{o}$ is chosen for the optimal transport coupling.} be the optimal coupling between $q^1_0$ and $q^2_0$.  Denote the push-forward of $\tilde{o}_0$ as $\tilde{o}_t^\# := [\phi_t]_\#\tilde{o}_0$. Then:

\begin{align}
\frac{d}{dt} \int_{\mathbb{R}^d \times \mathbb{R}^d} \|x-y\|^2 d\tilde{o}_t^\#(x,y)
&=\frac{d}{dt} \int_{\mathbb{R}^d \times \mathbb{R}^d} \|\phi_t(x)-\phi_t(y)\|^2 d\tilde{o}_0(x,y) \\
&=\int_{\mathbb{R}^d \times \mathbb{R}^d}  \frac{d}{dt} 
 \|\phi_t(x)-\phi_t(y)\|^2 d\tilde{o}_0(x,y) \\
&=\int_{\mathbb{R}^d \times \mathbb{R}^d}   
 2(\phi_t(x)-\phi_t(y))\cdot(\frac{d}{dt}\phi_t(x)-\frac{d}{dt}\phi_t(y)) d\tilde{o}_0(x,y) \\
&=2\int_{\mathbb{R}^d \times \mathbb{R}^d}   
 (\phi_t(x)-\phi_t(y))\cdot(u(t,\phi_t(x))-u(t,\phi_t(y))) d\tilde{o}_0(x,y) \\
&\le 2L \cdot \int_{\mathbb{R}^d \times \mathbb{R}^d}   
 \|\phi_t(x)-\phi_t(y)\|^2 d\tilde{o}_0(x,y) \\
&= 2L \cdot \int_{\mathbb{R}^d \times \mathbb{R}^d}   
\|x-y\|^2 d\tilde{o}_t^\#(x,y), 
\label{eq:wasserstein_derivative}
\end{align}

where in $(5)$ we changed variables, in $(6)$ we used Leibniz's rule and in $(8)$ the Lipshitz constraint.

Using Grönwall's inequality, we have:
\begin{align}
\int_{\mathbb{R}^d \times \mathbb{R}^d}   
\|x-y\|^2 d\tilde{o}_t^\#(x,y) &\leq \int_{\mathbb{R}^d \times \mathbb{R}^d}   
\|x-y\|^2 d\tilde{o}_0(x,y) \exp\{2Lt\}  \\
&=W_2^2(q^1_0,q^2_0)\exp\{2Lt\} .
\end{align}

By the definition of Wasserstein distance:
\begin{align}
W_2^2(q^1_t,q^2_t) \le \int_{\mathbb{R}^d \times \mathbb{R}^d}   
\|x-y\|^2 d\tilde{o}_t^\#(x,y).
\end{align}
Substituting the bound:
\begin{align}
W_2^2(q^1_t,q^2_t) \le W_2^2(q^1_0,q^2_0)\exp\{2Lt\}. 
\end{align}
Taking the square root of both sides yields the result:
\[
W_2(q^1_t, q^2_t) \leq W_2(q^1_0, q^2_0) \exp\{Lt\}.
\]

\end{proof}

\begin{theorem} \label{thm:trajectory_divergence_reductions_bound_prob}[Thm.\ref{thm:informal_trajectory_divergence_reductions_bound_prob} in the main paper]
Let $\mathcal{P}_2(\mathbb{R}^d)$ be the space of probability densities on $\mathbb{R}^d$ with finite second moments. Let $u$ be defined as in Lemma.~\ref{lemma:forward_backward_equality}. 
Let $\mu_t$ and $\nu_t$ 
be two time-dependent probability density functions in $\mathbb{R}^d$ satisfying the continuity equation Eq.~(\ref{eq:continuity_equation}) with initial densities $\mu_0$, $\nu_0$ and terminal densities $\mu_1$, $\nu_1$.
Suppose that the initial 2-Wasserstein distance between them is $d_0 = W_2(\mu_0, \nu_0)$. The time interval $[0,1]$ is discretized into $N$ equal steps: $t_n = n/N$, where $n = 0, 1, ..., N$. 
At each time step $t_n$, a time-dependent transport map $h:[0,1]\times \mathcal{P}_2(\mathbb{R}^d) \rightarrow \mathcal{P}_2(\mathbb{R}^d)$ is applied to $\nu_{t_n}$. After applying $h$, the flow function at time $t_{n+1}$ continues from the probability density of $h(t_n, \nu_{t_n})$.
We consider two variations for $h$:

\begin{enumerate}[label=\Alph*]
    \item If $h$ reduces the 2-Wasserstein distance by a constant amount $\epsilon_n > 0$ at each step: 
    $W_2(\mu_{t_n}, h(t_n, \nu_{t_n})) = W_2(\mu_{t_n}, \nu_{t_n}) - \epsilon_n$.
    Then, the following bound hold for $n = 0, 1, ..., N$: 
       $ W_2(\mu_{t_n}, \nu_{t_n}) \leq d_0 \exp\{L \cdot t_n\} - \sum_{i=0}^{n-1} {\epsilon_i \exp\{L(t_n - t_i)\}}$,

    \item If $h$ reduces the 2-Wasserstein distance to a proportion of the current distance:
    $W_2(\mu_{t_n}, h(t_n, \nu_{t_n})) = (1-\epsilon_{n}) \cdot W_2(\mu_{t_n}, \nu_{t_n})$ 
    where $0 < \epsilon_n < 1$ for all $n$.
    Then, the following bound hold for $n = 0, 1, ..., N$:
    $W_2(\mu_{t_n}, \nu_{t_n}) \leq d_0 \cdot \prod_{i=0}^{n-1} {(1-\epsilon_i)} \cdot \exp\{L\cdot t_n\}$. 

\end{enumerate}
\end{theorem}

\begin{proof}[Proof of Theorem \ref{thm:trajectory_divergence_reductions_bound_prob}] \label{proof:trajectory_divergence_reductions_bound_prob}

We will prove parts A and B of the theorem separately.

\textbf{Part A:}

Let $d_n = W_2(\mu_{t_n}, \nu_{t_n})$. We will prove by induction that:

\[
d_n \leq d_0 \exp\{L \cdot t_n\} - \sum_{i=0}^{n-1} {\epsilon_i \exp\{L(t_n - t_i)\}}.
\]

Base case ($n=0$): Trivially true as $d_0 = W_2(\mu_0, \nu_0)$.

Inductive step: Assume the inequality holds for $n$. We'll prove it for $n+1$.

After applying $h$ at $t_n$, we have:
\[
W_2(\mu_{t_n}, h(t_n, \nu_{t_n})) = d_n - \epsilon_n.
\]

By Lemma.~\ref{lemma:bound_distance_t} over the interval $[t_n, t_{n+1}]$, we have:
\[
d_{n+1} \leq (d_n - \epsilon_n) \exp\{L(t_{n+1} - t_n)\}.
\]

Substituting the inductive hypothesis:

\begin{align*}
d_{n+1} &\leq \left(d_0 \exp\{L \cdot t_n\} - \sum_{i=0}^{n-1} {\epsilon_i \exp\{L(t_n - t_i)\}} - \epsilon_n\right) \exp\{L(t_{n+1} - t_n)\} \\
&= d_0 \exp\{L \cdot t_{n+1}\} - \sum_{i=0}^{n-1} {\epsilon_i \exp\{L(t_{n+1} - t_i)\}} - \epsilon_n \exp\{L(t_{n+1} - t_n)\} \\
&= d_0 \exp\{L \cdot t_{n+1}\} - \sum_{i=0}^{n} {\epsilon_i \exp\{L(t_{n+1} - t_i)\}}.
\end{align*}

This completes the induction. The final bound at $t=1$ follows by setting $n=N$.

\textbf{Part B:}

Again, let $d_n = W_2(\mu_{t_n}, \nu_{t_n})$. We will prove by induction that:

\[
d_n \leq d_0 \cdot \prod_{i=0}^{n-1} {(1-\epsilon_i) \cdot \exp\{L\cdot n/N\}}.
\]

Base case ($n=0$): Trivially true.

Inductive step: Assume the inequality holds for $n$. We'll prove it for $n+1$.

After applying $h$ at $t_n$, we have:
\[
W_2(\mu_{t_n}, h(t_n, \nu_{t_n})) = (1-\epsilon_n) \cdot d_n.
\]

By Lemma.~\ref{lemma:bound_distance_t}, over the interval $[t_n, t_{n+1}]$ we have:
\[
d_{n+1} \leq (1-\epsilon_n) \cdot d_n \cdot \exp\{L(t_{n+1} - t_n)\} = (1-\epsilon_n) \cdot d_n \cdot \exp\{L/N\}.
\]

Substituting the inductive hypothesis:

\begin{align*}
d_{n+1} &\leq (1-\epsilon_n) \cdot \exp\{L/N\} \cdot d_0 \cdot \prod_{i=0}^{n-1} {(1-\epsilon_i) \cdot \exp\{L\cdot n/N\}} \\
&= d_0 \cdot \prod_{i=0}^{n} {(1-\epsilon_i) \cdot \exp\{L\cdot (n+1)/N\}}.
\end{align*}

This completes the induction. The final bound at $t=1$ follows by setting $n=N$.

\end{proof}

\textbf{Score Wasserstein Bound:}\label{appendix:score_wasserstein}

The score-based models losses are:
\begin{align*}
    L_{SM}(\theta; \lambda) &:= \frac{1}{2} \int_0^{T_s} \lambda(\tau) \mathbb{E}_{p_{\tau}} [\|\nabla \log p_{\tau}(x) - s_\theta(x, \tau)\|^2] d\tau, \\
    L_{DSM}(\theta, \lambda) &:= \frac{1}{2} \int_0^{T_s} \lambda(\tau) \mathbb{E}_{p_0(x(0))p_{0\tau}(x|x(0))}[\|s_\theta(x, \tau) - \nabla_x \log p_{0\tau}(x|x(0))\|^2] d\tau,
\end{align*}

where $\lambda:[0,T_s] \rightarrow (0, \infty)$ is a positive weighting function.  The score model is typically trained using the widely adopted loss function known as $L_{DSM}$.

\begin{definition}\label{definition:noisy_marginals}
    \textit{Noisy Backward Marginal}: Let $p^b_{n,\sigma}$ be the noisy $n^{th}$ backward marginal. Then, $p^b_{n,\sigma}(\tilde{y}_n|y_n)=\mathcal{N}(y_n,\sigma^2 I),\quad y\sim p^b_n$.
\end{definition}

\begin{theorem}[\cite{kwon2022score}]\label{theorem:wasserstein_bound}
    If $p_{0\tau}$ satisfies:
    \begin{align}
        Var[\mathbb{E}[(\nabla_x \log{p_{0\tau}(x|x(0)))^\top|x(0)}]]=0,
    \end{align}
    then we have:
    \begin{align}
        L_{SM} \leq L_{DSM} \quad \text{and} \quad W_2(p_0, q_0) \leq \sqrt{2\left(\int_0^{T_s} g(\tau)^2 I(\tau)^2 \, d\tau\right)L_{DSM}} + I(T_s)W_2(p_{T_s}, q_{T_s}),
    \end{align}
\end{theorem}

were $I(\tau):= \exp\{\int_0^\tau (L_f(r)+L_s(r)g(r)^2) dr\}$. Additionally, $L_f$ and $L_s$ are defined as follows:

\begin{enumerate}[label=(A\arabic*)]
    \item The drift coefficient $f : \mathbb{R}^d \times [0,T_s] \to \mathbb{R}^d$ is Lipschitz continuous in the space variable $x$: there exists a positive constant $L_f(\tau) \in (0, \infty)$, depending on $\tau \in [0,T_s]$, such that for all $x, y \in \mathbb{R}^d$
    \begin{align}
        \|f(x,\tau) - f(y,\tau)\| \leq L_f(\tau)\|x - y\|.
    \end{align}
    
    \item $s_\psi : \mathbb{R}^d \times [0,T_s] \to \mathbb{R}^d$ satisfies the one-sided Lipschitz condition [14, Definition 2.1]: there exists a constant $L_s(\tau) \in \mathbb{R}$, depending on $\tau \in [0,T_s]$, satisfying for all $x, y \in \mathbb{R}^d$
    \begin{align}
        (s_\psi(x,\tau) - s_\psi(y,\tau)) \cdot (x - y) \leq L_s(\tau)\|x - y\|^2.
    \end{align}
\end{enumerate}

For more details, see \cite{kwon2022score}.
In our case $p_0=p^b_{n}$ and $q_{T_s}=p^f_{n}$, the $n^{th}$ backward and forward marginals respectively.  We assume the forward marginals are a "noisy" version of the backward marginals (Def.~\ref{definition:noisy_marginals}) with a small $\sigma$.  Thus, $p_{T_s}=q_{T_s}$ and $W_2(p_{T_s}, q_{T_s})=0$ the first time the score model is applied, (the starting point is on the forward marginal). Consequently, as the $L_{DSM}$ loss converges to zero, so is the bound on the Wasserstein distance. 

\textbf{Conditions:} 
$Var[\mathbb{E}[(\nabla_x \log{p_{0\tau}(x|x(0)))^\top|x(0)}]]=0$ under the following sufficient conditions: (1) Lipschitz continuity of the drift function for the forward diffusion process, and (2) boundedness of the noise schedule. Our scenario satisfies these conditions as follows:
\begin{itemize}
    \item The drift function $f$ is zero, which is Lipschitz continuous.
    \item The noise schedule $g$ is between $\sigma_{min}$ and $\sigma_{max}$, thus it is bounded above and below.
\end{itemize}

Therfore, we meet these sufficient conditions and adopt the additional assumptions from \citet{kwon2022score}, enabling us to apply Thm.~\ref{theorem:wasserstein_bound}.

\begin{lemma}%
    Let $u: [0, 1] \times \mathbb{R}^d \rightarrow \mathbb{R}^d$ be a continuous time-dependent vector field, satisfying the Lipschitz condition: for any $t \in [0, 1]$ and $x, y \in \mathbb{R}^d$, 
    $\|u_t(x) - u_t(y)\| \leq L \|x - y\|$. Let $p^b_n$ and $p^f_n$ represent the $n^{th}$ backward and forward marginals respectively, where the corresponding \textbf{marginals} continuous time is $t_n=n/N$. Let $s_\psi$ denote the trained score model. Then, applying $s_\psi$ from $p^f_n$ to $p^b_n$ 
     as described in Sec.~\ref{section:score_models} lowers the final bound on $W_2(p^b_N,p^f_N)$:
    \[W_2(p^b_N,p^f_N) \le \left(\sqrt{2\left(\int_0^{T_s} g(\tau)^2I(\tau)^2d\tau\right)L_{DSM}^\psi}\right)\exp\{L(1-t_n)\}.\] 
\end{lemma}

\begin{proof}[Proof of Lemma \ref{lemma:one_step_score}]\label{proof:one_step_score}
    Denote as $d_n=W_2(p^b_{n},p^f_{n})$.
    After applying the score models $s_\psi$:
    \begin{align}
        \hat{d}_{n,\psi} = W_2(p^b_{n}, \hat{p}^b_{n}) \leq \sqrt{2\left(\int_0^{T_s} g(\tau)^2I(\tau)^2d\tau\right)L_{DSM}^\psi} + I(T_s)W_2(p^b_{n,\sigma}, p^f_{n}), 
    \end{align}
    where $\hat{p}^b_{n}$ is the score model's approximation and $0,T_s$ are the integration times of the score model and ${T_s} \rightarrow \infty$. 
    According to Lemma.~\ref{lemma:bound_distance_t}:
    \begin{align}
        W_2(p^b_N,p^f_N)&\le \hat{d}_{n,\psi} \exp\{L(1-t_n)\}, \\
        W_2(p^b_N,p^f_N)&\le W_2(p^b_n, p^f_n) \exp\{L(1-t_n)\}=W_2(p^b_n, p^b_{n,\sigma}) \exp\{L(1-t_n)\},
    \end{align}
    
   where the second inequality is the bound without applying the correction step.  
   
    The final bound can be obtained by substituting the bound on $\hat{d}_{n,\psi}$:
    
    \begin{align}
        W_2(p^b_N,p^f_N) &\le \left(\sqrt{2\left(\int_0^{T_s} g(\tau)^2I(\tau)^2d\tau\right)L_{DSM}^\psi} + I(T_s)W_2(p^b_{n,\sigma}, p^f_{n})\right)\exp\{L(1-t_n)\}.
    \end{align}

    Assuming the score model is trained, $J^\psi_{DSM}$ is negligible: $\sqrt{2\left(\int_0^{T_s} g(\tau)^2I(\tau)^2d\tau \right)L_{DSM}^\psi}<W_2(p^b_n, p^b_{n,\sigma})$. Since the forward marginals are a noisy version of the backward marginals $W_2(p^f_n,p^b_{n,\sigma})=0$. In total, the final bound after applying a correction step is lower than the bound when applying no correction steps:
    \begin{align}
        \left(\sqrt{2\left(\int_0^{T_s} g(\tau)^2I(\tau)^2d\tau\right)L_{DSM}^\psi}\right)\exp\{L(1-t_n)\} < W_2(p^b_n, p^b_{n,\sigma}) \exp\{L(1-t_n)\}.
    \end{align}

\end{proof}

\begin{theorem}\label{thm:correction_score_multiple}
    Let $u: [0, 1] \times \mathbb{R}^d \rightarrow \mathbb{R}^d$ be a continuous time-dependent vector field, satisfying the Lipschitz condition: for any $t \in [0, 1]$ and $x, y \in \mathbb{R}^d$, 
    $\|u_t(x) - u_t(y)\| \leq L \cdot \|x - y\|$.  
    Denote $l^{\psi,n}=\sqrt{2\left(\int_0^{T_s} g(\tau)^2I(\tau)^2d\tau\right)L_{DSM}^{\psi,n}}$, where $L_{DSM}^{\psi,n}$ refers to the loss of $s_\psi$ on the $n^{th}$ marginal.
    Following Lemma.~\ref{lemma:one_step_score} and assuming that $W_2(p^b_{n,\sigma_n},q_{T_s,n}) \le W_2(p^b_{n},q_{T_s,n})$ for every $n$ where $q_{T_s,n}$ is the initial distribution of the $n^{th}$ correction of the score model, then the application of $s_\psi$ to multiple marginals upper bounds the final Wasserstein distance. This can be expressed as:
    \[W_2(p^b_N,p^f_N) \le \sum_{i=0}^{N-1} l^{\psi,i} \exp\{(N-i) \cdot L/N\} I^{N-1-i}(T_s) + d_0 I^{N}(T_s) \exp\{L\}.\]
\end{theorem}

Under the assumption that the forward marginals are a noisy version of the backward marginals when $q_{T_s,n}=p^f_n$ the inequality: $W_2(p^b_{n,\sigma_n},p^f_n) \le W_2(p^b_{n},p^f_n)$ is trivial.  After applying a correction step and continuing the flow, the trajectory at the next time-step lies between the forward and backward marginal paths, rather than strictly on the forward marginals path ($q_{T_s,n} \neq p^f_n$). This inequality holds true provided that $\sigma_n$ decreases at a rate corresponding to this intermediate positioning.  

\begin{corollary}\label{corollary:correction_score_multiple}
Assuming the score model is well-trained, $l^{\psi,i}$ becomes negligible. In our case, the drift is 0, resulting in a negligible $L_f(r)$.
    Furthermore, \cite{kwon2022score} showed that with a DDPM schedule and $q_{T_s,n}=\mathcal{N}(0,I)$, the one-sided Lipschitz constant is $\lim_{\tau \rightarrow \infty}{\sigma^2 L_s(\tau)}=-1$. This can be generalized to a VE schedule. We employ such a schedule, but lack a closed form for $q_{T_s,n}$. Assuming $\lim_{\tau \rightarrow \infty}{\sigma^2 L_s(\tau)}=-1$ holds in our case as well, $\exists \tau'$ such that $\forall \tau > \tau'$ $I(\tau)<1$, specifically $\lim_{\tau \rightarrow \infty}{I(\tau)}<1$.  Applying more correction score steps decreases the upper bound of the final Wasserstein distance by $I(T_s)$, similarly to the multiplicative case (B) in Thm.~\ref{thm:trajectory_divergence_reductions_bound_prob}.

\end{corollary}

\begin{proof}[Proof of Theorem \ref{thm:correction_score_multiple}] \label{proof:correction_score_multiple}
Denote as $d_n=W_2(p^b_{n},q_{T_s,n})$. We will prove by induction that:

\begin{align}
d_n \leq \sum_{i=0}^{n-1} l^{\psi,i} \exp\{(n-i) \cdot L/N\} I^{n-i-1}(T_s) + d_0 I^{n}(T_s) \exp\{n \cdot L/ N\}.
\end{align}

Base case ($n=0$): Trivially true.

Inductive step: Assume the inequality holds for $n$. We'll prove it for $n+1$.

By Lemma.~\ref{lemma:one_step_score}, after applying the score function:
\begin{align}
    \hat{d}_{n,\psi} &= W_2(p^b_{n}, \hat{p}^b_{n}) \leq \sqrt{2\left(\int_0^{T_s} g(\tau)^2I(\tau)^2d\tau\right)L_{DSM}^{\psi,n}} + I(T_s)W_2(p^b_{n,\sigma_n}, q_{T_s,n}) \\
    &\le \sqrt{2\left(\int_0^{T_s} g(\tau)^2I(\tau)^2d\tau\right)L_{DSM}^{\psi,n}} + I(T_s)W_2(p^b_{n}, q_{T_s,n}) \\
    &= \sqrt{2\left(\int_0^{T_s} g(\tau)^2I(\tau)^2d\tau\right)L_{DSM}^{\psi,n}} + I(T_s)d_n,
\end{align}

in the second inequality we used the assumption that $W_2(p^b_{n,\sigma_n},q_{T_s,n}) \le W_2(p^b_{n},q_{T_s,n})$. By Lemma.~\ref{lemma:bound_distance_t} over the interval $[t_n, t_{n+1}]$, we have:
\begin{align}
    d_{n+1} &\le \hat{d}_{n,\psi} \exp\{L(t_{n+1}-t_n)\} \\
    &\le \left(\sqrt{2\left(\int_0^{T_s} g(\tau)^2I(\tau)^2d\tau\right)L_{DSM}^{\psi,n}} + I(T_s)d_n \right) \exp\{L(t_{n+1}-t_n)\} \\
    &= \left(\sqrt{2\left(\int_0^{T_s} g(\tau)^2I(\tau)^2d\tau\right)L_{DSM}^{\psi,n}}\right)\exp\{L/N\} +d_n I(T_s)\exp\{L/N\},
\end{align}

Substituting the inductive hypothesis:

\begin{align}
&d_{n+1} \le l^{\psi,n} \exp\{L/N\} + d_n I(T_s) \exp\{L/N\} \\
&\le l^{\psi,n} \exp\{L/N\} \\ 
&+ \left( \sum_{i=0}^{n-1} l^{\psi,i} \exp\{(n-i) \cdot L/N\} I^{n-i-1}(T_s) + d_0 I^{n}(T_s) \exp\{n \cdot L/ N\}\right)I(T_s) \exp\{L/N\}  \\
&= \sum_{i=0}^{n} l^{\psi,i} \exp\{(n-i+1) \cdot L/N\} I^{n-i}(T_s) + d_0 I^{n+1}(T_s) \exp\{(n+1) \cdot L/ N\}.
\end{align}

This completes the induction. The final bound at $t=1$ follows by setting $n=N$.

\end{proof}

\section{Marginal Matching Inference}

Algorithm~\ref{alg:our_predictor_corrector} presents our proposed approach. 
The correction steps using $h_\psi$ are taken before each step of the flow model $v_\theta$ and at the end.
For clarity, the algorithm assumes Euler integration for sampling $v_\theta$ and single correction steps, though multiple corrections per time-step are possible. Adding Gaussian perturbation before each correction step, except the last one, improves performance. 

\begin{algorithm}[th]
    \centering
    \caption{Marginal Matching}\label{alg:our_predictor_corrector}
    \begin{algorithmic}[1]
    \REQUIRE {flow model $v_\theta$, correction model $h_\psi$, number of iterations $N$, step size of corrector steps $\{\alpha_i\}_{i=0}^N$, scale added noise $\{\beta_i\}_{i=0}^N$}
    \STATE $x_0 \sim \mathcal{N}(0,I)$
    \FOR {$n=0,\dots,N-1$}
    \STATE $\epsilon \sim \mathcal{N}(0,I)$
    \STATE $\tilde{x}_{t_n} = x_{t_n} + \beta_n\cdot \epsilon$
    \STATE $x_{t_n} = x_{t_n} + \alpha_n \cdot h_\psi(t_n, \tilde{x}_{t_n})$ \COMMENT{Correction}
    \STATE $x_{t_{n+1}}=x_{t_n} + \frac{1}{N} v_\theta(t_n, x_{t_n})$ \COMMENT{Flow}
    \ENDFOR
    \STATE $x_{t_N} = x_{t_N} + \alpha_N\cdot h_\psi(t_N, x_{t_N})$ \COMMENT{Correction}
    \STATE \textbf{return} $x_{t_N}$
    \end{algorithmic}
\end{algorithm}

\vspace{8ex}

\section{Practical Considerations}\label{appendix:practical_considerations}

\subsection{Backward Marginals Generation}\label{appendix:backward_generation}

Once a flow model is defined, we can generate samples from the backward marginal distributions using any ODE solver to integrate from time t=1 to t=0. We present two example implementations in Algorithm~\ref{alg:rk4_backward} and Algorithm~\ref{alg:euler_backward}, using RK4 and Euler integration methods respectively. The intermediate states at each time-step correspond to samples from the backward marginal distributions at those times (e.g., at t=1 we sample from $p_1^b$, and at t=0 we sample from $p_0^b$).

\begin{algorithm}
\caption{Backward Marginal Generation with RK4 Integration}
\begin{algorithmic}[1]
\REQUIRE Initial condition $y(1)$, number of steps $N$, flow model $v_\theta$
\STATE $h \gets -1/N$ \COMMENT{Negative step size for backward integration}
\STATE $t \gets 1$ \COMMENT{Start time}
\STATE $y \gets y(1)$ \COMMENT{Initial value}

\FOR{$i = 1$ to $N$}
    \STATE $k_1 \gets h \cdot v_\theta(t, y)$
    \STATE $k_2 \gets h \cdot v_\theta(t + \frac{h}{2}, y + \frac{k_1}{2})$
    \STATE $k_3 \gets h \cdot v_\theta(t + \frac{h}{2}, y + \frac{k_2}{2})$
    \STATE $k_4 \gets h \cdot v_\theta(t + h, y + k_3)$
    
    \STATE $y \gets y + \frac{1}{6}(k_1 + 2k_2 + 2k_3 + k_4)$
    \STATE $t \gets t + h$
\ENDFOR

\STATE \textbf{return} $y$ \COMMENT{Solution at t=0}
\end{algorithmic}\label{alg:rk4_backward}
\end{algorithm}

\begin{algorithm}
\caption{Backward Marginal Generation with Euler Integration}
\begin{algorithmic}[1]
\REQUIRE Initial condition $y(1)$, number of steps $N$, flow model $v_\theta$
\STATE $h \gets -1/N$ \COMMENT{Negative step size for backward integration}
\STATE $t \gets 1$ \COMMENT{Start time}
\STATE $y \gets y(1)$ \COMMENT{Initial value}

\FOR{$i = 1$ to $N$}
    \STATE $y \gets y + h \cdot v_\theta(t, y)$ \COMMENT{Euler step}
    \STATE $t \gets t + h$
\ENDFOR

\STATE \textbf{return} $y$ \COMMENT{Solution at t=0}
\end{algorithmic}\label{alg:euler_backward}
\end{algorithm}

\subsection{Practical Considerations of $h_\psi$} 
When designing the learning objective for $h_\psi$ an important issue should be taken into consideration. The pre-trained flow model has established a matching between each distribution on the forward trajectory $p^f_{t_n}$ and between each distribution on the backward trajectory $p^b_{t_n}$ across different time-steps. Additionally, there exists a matching between $\hat{\pi}_1$ and $\pi_1$. Through transitivity, this implies a pairing between $p^f_{t_n}$ and $p^b_{t_n}$ for every time step ${t_n}$. During the training process, samples from $p^f_{t_n}$ and $p^b_{t_n}$ are accessible. 
Models that try to establish pairing between these distributions will yield sub-optimal results, since the correct pairing is unknown.
Thus, we opt for different generative models such as score and classifier.

\textbf{Alternative Correction Models:} Alternative correction models include the more sophisticated score models for traversing between forward and backward marginals, provided the noise level for the forward marginal is known. A supplementary model can be trained to predict the noise level, thereby assisting the score model.
Additionally, Energy-Based Models (EBMs) \citep{du2021improved}, could also be a good fit to explore the alignment of the sampling trajectory with the backward trajectory.

\subsection{Approximations}\label{appendix:approximations}
\subsubsection{Interpolating $p_0^b$ and $\pi_1$} \label{approximating_marginals} 
The correction model $h_\psi(t,x_t)$ is trained on different time-step marginals. The marginals are generated and stored in advance to avoid running the ODE-solver multiple times while training $h_\psi(t,x_t)$.Assuming the flow model's trajectories are sufficiently straight, we can save this extra storage by approximating the backward trajectory through interpolation between $p_0^b$, the approximate source distribution, and $p_1^b$, the target distribution: $\hat{p}^b_{t_n}=t_n\cdot p^b_1 + (1-t_n) \cdot p^b_0$. This reduces storage requirements to just the data and approximate source distribution, making the dataset size twice rather than $N$ times the original size.
Table.~\ref{table:parallel_approximation_marginals} presents the results of the interpolated backward trajectory on CIFAR-10. The score model's performance remains steady, while the classifier exhibits degradation in its performance.
Table.~\ref{table:imagenet_approximation_marginals} compares the results of a score correction model on ImageNet-64 trained on the backward marginals $p^b_t$ versus the approximated marginals. The model trained on approximated marginals shows improved results, albeit not as significantly as the non-approximated version.

\subsubsection{Parallel Sampling} \label{appendix:parallel_sampling}

The corrected inference process, as described in Algorithm~\ref{alg:our_predictor_corrector}, introduces additional sampling time for each correction step. 
To reduce the overhead, we propose a correction algorithm that is parallelizable in principle, by running the correction  and the flow model in parallel. The correction and the flow model both operate on the same input at the same time, then the correction $h_\psi$ and the flow model's $v_\theta$ direction are summed, approximating the full correction process. 
While this is an approximation, the effectiveness of this approximation relies on the assumption that $h_\psi$ and $v_\theta$ do not change too rapidly.

\begin{figure}[t]
  \begin{center}
    \includegraphics[width=0.3\textwidth]{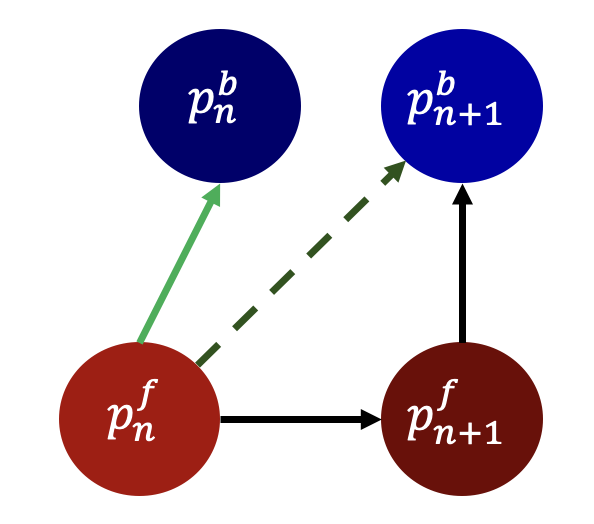}
  \end{center}
  \caption{\textbf{Parallel Sampling} The black arrows denote the path of first flowing with $v_\theta$ and only then taking a correction step with $h_\psi$.  In contrast, the dashed green arrow shows the approximation obtained by summing the direction of the flow model $v_\theta$ (black arrow), and the correction model $h_\psi$ (green arrow) on the current step.}
  \label{fig:parallel_sampling}
\end{figure}

Fig.~\ref{fig:parallel_sampling} presents the approximation of parallel sampling the correction and flow model.
The algorithm for parallel sampling is presented in Alg.~\ref{alg:parallel_predictor_corrector}.  For clarity, the algorithm assumes Euler integration as the ODE-solver. To maintain computational efficiency, a single correction step is implemented, ensuring that the total computation time remains equivalent to that of the uncorrected method.

This algorithm is parallel in principle. Cuda has a queue, where the CPU sends tasks to be run on the GPU. The GPU  may execute tasks in parallel that are independent of one another, such as our parallel sampling. The CPU may wait when the queue is full or during synchronization events, (i.e. item(), synchronize(), etc). Given a powerful enough Cuda machine, the models could be run in parallel on different GPUs or using advanced parallelism techniques. Parallel calculation on separate devices could be useful when the correction calculation takes more time than the overhead of moving between devices, as is the case for large models. The extent of parallel processing viability will depend on the specific hardware and infrastructure available.

\begin{algorithm}[hb]
    \centering
    \caption{Parallel Marginal Matching}\label{alg:parallel_predictor_corrector}
    \begin{algorithmic}[1]
    \REQUIRE {flow model $v_\theta$, correction model $h_\psi$, number of iterations $N$, step size of corrector steps $\{\alpha_i\}_{i=0}^{N-1}$, scale added noise $\{\beta_i\}_{i=0}^{N-1}$}
    \STATE $x_0 \sim \mathcal{N}(0,I)$
    \FOR {$n=0,\dots,N-1$}
    \STATE $\epsilon \sim \mathcal{N}(0,I)$
    \STATE $\tilde{x}_{t_n} = x_{t_n} + \beta_n\cdot \epsilon$
    \STATE $h_{t_n} = h_\psi(t_n, \tilde{x}_{t_n})$ \COMMENT{Correction}
    \STATE $v_{t_n} = v_\theta(t_n, x_{t_n})$ \COMMENT{Flow}
    \STATE $x_{t_{n+1}} = x_{t_n} + \frac{1}{N} v_{t_n} + \alpha_n \cdot h_{t_n}$
    \ENDFOR
    \STATE \textbf{return} $x_{t_N}$
    \end{algorithmic}
\end{algorithm}

Table.~\ref{table:parallel_approximation_marginals} presents the complete results on CIFAR-10 of applying the correction models (score and classifier) in parallel with the flow model, (to save computation time), and with an approximation of the backward marginals (to save storage during training).  

\begin{table}[ht]%
\centering
\begin{tabular}{ l |  l l  c c }

&&&OT-CFM &I-CFM \\ 
  \textit{Correction Model} & NFE $\downarrow$ & C-NFE $\downarrow$ & FID $\downarrow$ & FID $\downarrow$ \\
\bottomrule
  No Correction &40 & 0 & 4.34 &4.29\\
\bottomrule

\multirow{4}{9em}{Classifier} &42 &1 ~~\textit{($n=8$)}  & 3.57 &3.77 \\  
 & 46 &3 ~~\textit{($n=8,9,10$)}  & 3.48 &3.67 \\  
&50 &5 ~~\textit{($n=6,7,8,9,10$)}  & \textbf{3.47} & \textbf{3.62} \\
&62 &11 \textit{($\forall n$)}  & 3.48 &3.63 \\ 
 \midrule
 P-Classifier &40 & 1 ~~\textit{($n=7$)} & 3.65 & 3.75 \\
 &40 & 3 ~~\textit{($n=7,8,9$)} & 3.60 & 3.73 \\ 
&40 & 5 ~~\textit{($n=5,6,7,8,9$)} & \textbf{3.59} & \textbf{3.67}\\ 
&40 & 10 \textit{($n=0,1,2,3,4,5,6,7,8,9$)} & 3.74 & 3.80\\ 
 \midrule
 \multirow{4}{9em}{Classifier \\ {\footnotesize(Backward Marginals} {\footnotesize Interpolation)}} &42  &1 ~~\textit{($n=8$)} & 3.77 & 3.74\\  
&46 &3 ~~\textit{($n=8,9,10$)} &  3.72 & \textbf{3.62}\\  
&50 &5 ~~\textit{($n=6,7,8,9,10$)}  & \textbf{3.67} & 3.69 \\
 &62 &11 \textit{($\forall n$)}  & 3.68 & 3.67\\
\bottomrule
 \multirow{3}{9em}{Score} &41  &1 ~~\textit{($n=10$)}  & 3.45 & 3.47\\  
 &43 &3 ~~\textit{($n=0,5,10$)}  & 3.38 & 3.39\\  
&51 &11 \textit{($\forall n$)}  & \textbf{3.37} & \textbf{3.38}\\
 \midrule
 P-Score &40 & 1 ~~\textit{($n=9$)}& 4.29 & 4.27 \\
&40 & 3 ~~\textit{($n=0,5,9$)}&  \textbf{4.08} & \textbf{3.99}\\
&40 & 10 \textit{($n=0,1,2,3,4,5,6,7,8,9$)}& \textbf{4.08} & \textbf{3.99} \\
 \midrule
  \multirow{3}{8em}{Score \\{\footnotesize(Backward Marginals} {\footnotesize Interpolation)}} &41 &1 ~~\textit{($n=10$)}  & 3.46 &3.47 \\  
 &43 &3 ~~\textit{($n=0,5,10$)}  & \textbf{3.38} & \textbf{3.40}\\  
 &51 &11 \textit{($\forall n$)}  & 3.41 & 3.41\\
\bottomrule
\end{tabular}
\caption{
Comparison of two variants of classifier and score correction models on CIFAR-10.
First, a correction model that runs in parallel with ("P-") the flow model. 
Second, a correction model that learns an interpolation of the data and approximate source distribution. 
The base flow models are sampled with 10-step RK4 (40 NFE).
The classifier succeeds in improving the FID even with the parallel approximation, while the score shows less improvement. 
Training with interpolated backward marginals improves the FID score, though less than using their exact version.
} 
\label{table:parallel_approximation_marginals}
\end{table}

\begin{table}[ht]%
\centering
\begin{tabular}{ l |  l l  c }

  \textit{Correction Model} & NFE $\downarrow$ & C-NFE $\downarrow$ & FID $\downarrow$ \\
\bottomrule

 \multirow{3}{*}{Score} &21&1 \textit{($n=5$)}& 15.91\\
 &22&2 \textit{($n=0,5$)}& \textbf{15.53}\\
 &26&6 \textit{($n=0\dots5$)}& 15.57 \\
 \midrule

 \multirow{3}{9em}{Score \\ {\footnotesize(Backward Marginals} {\footnotesize Interpolation)}} &21 &1 \textit{($n=5$)}& 15.89\\
 &22&2 \textit{($n=0,5$)}& \textbf{15.62}\\
 &26&6 \textit{($n=0\dots5$)}& 15.69\\
\bottomrule
\end{tabular}
\caption{
Comparison of backward marginals $p^b_t$ and approximated marginals results on ImageNet-64 using OT-CFM as the base flow model, sampled with $5$-step RK4 ($20$ NFE). Both versions improve the original score, though the approximation's benefit is smaller.
}
\label{table:imagenet_approximation_marginals}
\end{table}

\subsection{Score Model Training}

Score models are generative AI systems that transform random noise into meaningful data through iterative denoising. At their core is the score function - the gradient of the log probability density - which guides samples toward higher likelihood regions of a target distribution. 

These models utilize varying noise levels ($\sigma$) during generation. At high noise levels, samples differ significantly from the target distribution, but the score function provides stable gradients far from the data manifold. At low noise levels, as samples approach the target distribution, the score function enables detailed refinement. During generation, the model progressively reduces noise levels while following each corresponding score function, establishing a path between random noise and complex data distributions.

By assuming the forward marginals are a noisy version of the backward marginals (with a small $\sigma$), a score model can be used to transverse between them.

While sophisticated approaches like annealing score models with time-varying $\sigma$ are widely used for generation, they proved unsuitable for our needs. This is due to our inference process that begins at a forward marginal (or an intermediate point), where we lack information about the current noise level, precluding the effective use of such models.

\subsection{Robust Classifier Training} 
\label{appendix:classifier_training}
\textbf{Adversarial Training (AT):} \citet{srinivas2021rethinking} demonstrated that the gradients of standard classifiers can be arbitrarily altered without impacting their cross-entropy loss or accuracy. Building on this insight, \citet{kawar2023enhancing} proposed using gradients from a \textit{robust} classifier for guidance, rather than those from a conventional classifier.
The gradients of robust classifiers are resistant to arbitrary manipulation as a results of the adversarial attack used in their loss computation is directly dependent on the model's gradients. An intriguing characteristic of robust classifiers is that their gradients have been shown to align well with human perception, as noted by \citet{tsipras2018robustness}. 
When robust classifiers are employed to guide $x$ towards a specific class c, they are anticipated to produce significant features that correspond well with the target class. As a result, the modifications applied to $x$ are likely to be visually convincing and aligned with human perception of the class characteristics.
Inspired by these works we enhance the classifier with Projected Gradient Descent (PGD) attack and Adversarial Training (AT) robustification method (\citet{madry2018towards}), with the PGD pseudo-algorithm detailed in Alg.~\ref{alg:Targeted_Projected_Gradient_Descent}. Our findings indicate that adversarial training stabilizes gradients and enhances their meaningfulness.

\begin{algorithm}%
    \centering
    \caption{Targeted Projected Gradient Descent}
    \begin{algorithmic}[1]
    \REQUIRE {classifier $f_\phi$, input $x$, class $c$, number of iterations $N$, step size $\alpha$, radius $\epsilon$, loss function $\ell$}, 
    \STATE $\delta_0 = 0$
    \FOR {$n=0,\dots,N$}
        \STATE $\delta_{n+1}=\Pi_\epsilon(\delta_n-\alpha\nabla_\delta \ell(f_\phi(x+\delta_n), c))$
    \ENDFOR
    \STATE $x_{ADV} = x + \delta_{N}$
    \STATE Return $x_{ADV}$
    \end{algorithmic}
\label{alg:Targeted_Projected_Gradient_Descent}
\end{algorithm}

Where $\Pi_\epsilon$ is a projection operator that keeps $\delta$'s norm below $\epsilon$ so the adversarial example will not stray too far from the input; and the loss function in our case is cross entropy.

\textbf{Gradient Alignment (GA):} Inspired by \citet{yadin2024classification} and \citet{song2021scorebased}, we incorporate a cosine-similarity loss term for the gradient. This aims to align the classifier's gradient direction with the backward marginals when in close proximity. We artificially move away from the backward marginals by introducing small Gaussian perturbations to sampled points, and training the classifier gradient to point towards the original samples by aligning them with the negative direction of the noise.

\begin{algorithm}%
    \centering
    \caption{Robust Classifier Training}\label{alg:classifier_training}
    \begin{algorithmic}[1]
    \REQUIRE {classifier $h_\psi$, forwards marginals $\{p_{t_n}(x)^f\}_{n=0}^N$, backward marginals $\{p_{t_n}(x)^b\}_{n=0}^N$, loss weights $\{\beta_i\}_{i=1}^3$, $\sigma$ scale of noise, adversarial step size $\alpha$, adversarial number of steps $K$, adversarial radius $\epsilon_{ADV}$}
    \REPEAT
    \STATE $\epsilon \sim \mathcal{N}(0,I)$
    \STATE $n\sim U(\{0,1,..,N\})$
    \STATE $s = \mathcal{U}(0,1) \cdot \sigma$
    \STATE $x_f\sim p_{t_n}^f , x_b\sim p_{t_n}^b$
    \STATE $\ell_{CE}=CE(h_\psi(x_f,t_n),0)+CE(h_\psi(x_b,t_n),1)$ 
    \STATE $c_n = \nabla_\psi h_\psi(x_b+s \cdot \epsilon,t_n)$
    \STATE $\ell_{CS}=cosine-similarity(c_n,-\epsilon)$
    \STATE $\ell_{ADV}=PGD(h_\psi, x_f, 0, K, \alpha, \epsilon_{ADV}, CE)+ PGD(h_\psi, x_b, 1, K, \alpha, \epsilon_{ADV}, CE)$
    \STATE $\ell=\beta_1 \cdot \ell_{CE}+\beta_2 
\cdot \ell_{CS}+\beta_3 \cdot \ell_{ADV}$
    \STATE Take gradient descent step on $\nabla_\psi\ell$
    \UNTIL {converged}
    \end{algorithmic}
\end{algorithm}

The class for the forward marginals is represented by $0$ and for the backward marginals by $1$.

\section{Extended Related Work}\label{appendix:extened_related_work}

\textbf{Different sampling trajectories} -- 
Previous work explored different sampling trajectories primarily by studying various source distributions. 
In Denoising Diffusion Probabilistic Models \citep{ho2020denoising,sohl2015deep}, various source distributions proved to decrease inference time while maintaining quality. \citet{lee2022priorgrad} 
proposed using a Gaussian prior distribution with parameters derived from data statistics. \citet{lyu2022accelerating,popov2021grad} trained a model for prior distribution using an encoder or VAE \citep{kingma2013auto} and incorporated it into the standard diffusion process by adding noise. Other studies have shown that the prior distribution need not be Gaussian \citet{bansal2024cold, heitz2023iterative}. Flow based models \citet{lipman2023flow,tong2024improving,liu2023flow}, a new family of generative models, generalized the mapping to include general source distributions.  

A work aligned with ours is that of \citet{xu2024normalizing} as they consider the learned errors throughout the whole sampling trajectory. They use a sequence of NF blocks to learn a sequential transformation from Gaussian noise to data, where each block represents a different time-step.
This requires training the blocks sequentially, as the learned error in each block is accounted for in the next block, slowing the training process. The sampling process initiates from Gaussian noise, where the model's error is not accounted for. In contrast, our work  differs in that we focus on correcting the inference-time trajectory rather than the training trajectory, and we enhance flow matching models instead of NF. 

\textbf{Training and inference mismatch} -- Previous work recognized training and inference mismatch in different settings.  
In score models (\citet{song2021scorebased,song2019generative}), the source distribution during inference is $\mathcal{N}(0,I)$, which is intended to approximate the "noisiest" training distribution ($p_T$). However, as these distributions are not identical, starting from $\mathcal{N}(0,I)$ will yield sub-optimal results. Two different solutions were proposed to minimize this gap as a pre-sampling procedure: \citet{franzese2023much} suggested an auxiliary model that learns the mapping from $\mathcal{N}(0,I)$ to $p_T$. Alternatively, \citet{pedrotti2024improved} proposed to use Langevin dynamics (\citet{welling2011bayesian}), leveraging the score function $\nabla_x\log{p_T(x)}$.

Even if the source distribution during training and inference is the same, practically, the actual distribution that reaches the data is different.
\citet{coeurdoux2023normalizingflowsamplinglangevin} leveraged the invertibility property of normalizing flow (NF) models (\citet{kingma2018glow,dinh2014nice}) to reach higher probability samples in $\hat{\pi}_1$, which are more likely to belong to $\pi_1$. They achieved this by employing MCMC algorithm, where the score is computed using the Jacobian of the inverse mapping. 
This discrepancy between the learned and actual source distribution is common in generative models like VAEs \citep{kingma2021variational}, which assume a Gaussian prior during inference. \citep{dai2018diagnosing} addressed this by learning an additional VAE model to predict the actual prior distribution, thereby removing the Gaussian assumption, which lead to improved results.

\textbf{Inverse diffusion} -- Previous work researched inversion of diffusion models, which are "approximately invertible" models, particularly in the context of image editing. DDIM inversion \citep{song2021denoising} method laid the groundwork for this approach, enabling the reversal of the diffusion process to obtain latent representations of images by utilizing a deterministic forward process. Building upon this foundation, \citet{wallace2023edict}  improved the efficiency of the inversion by training an encoder network to directly map images to their corresponding noise representations in the diffusion process. Furthermore, \citet{mokady2023null} enabled precise edits on real images through a text-guided approach, using a null-text optimization to find an optimal noise that, when denoised, produces the target image. \citet{zhang2023exact} improved the accuracy of the inversion process by combining forward and backward trajectories to minimize approximation errors, while \citet{pan2023effective} focused on enhancing both the speed and quality of image editing operations by iteratively refining the inverted latent code.

\section{Forward and Backward Trajectory Comparison}\label{appendix:forward_vs_backward}

\begin{figure}[H]
\centering
 \begin{subfigure}{0.23\textwidth}
     \includegraphics[width=\textwidth]{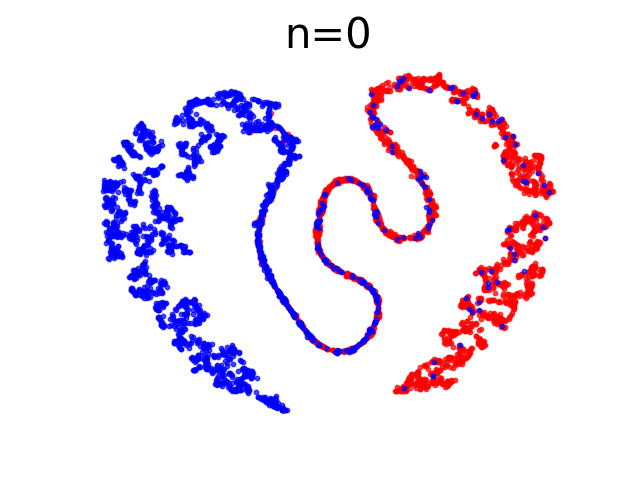}
 \end{subfigure}
 \begin{subfigure}{0.23\textwidth}
     \includegraphics[width=\textwidth]{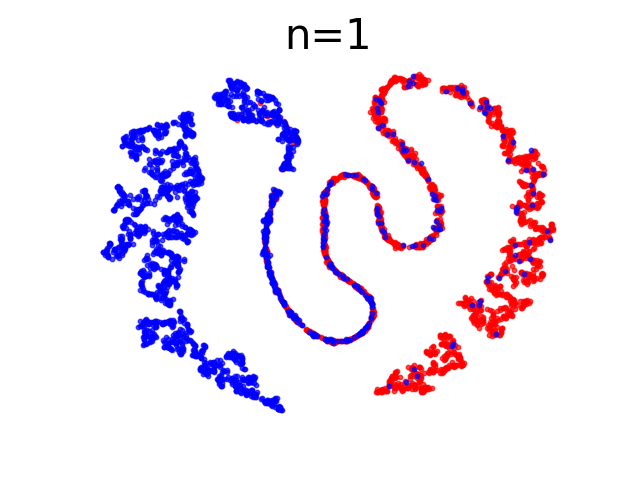}
 \end{subfigure}
 \begin{subfigure}{0.23\textwidth}
     \includegraphics[width=\textwidth]{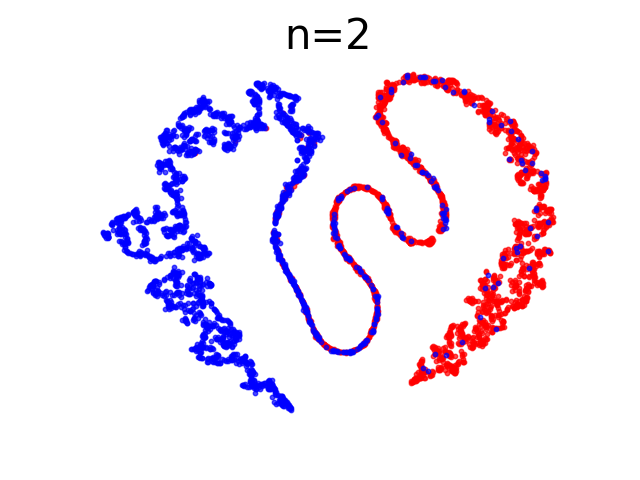}
 \end{subfigure}
 \begin{subfigure}{0.23\textwidth}
     \includegraphics[width=\textwidth]{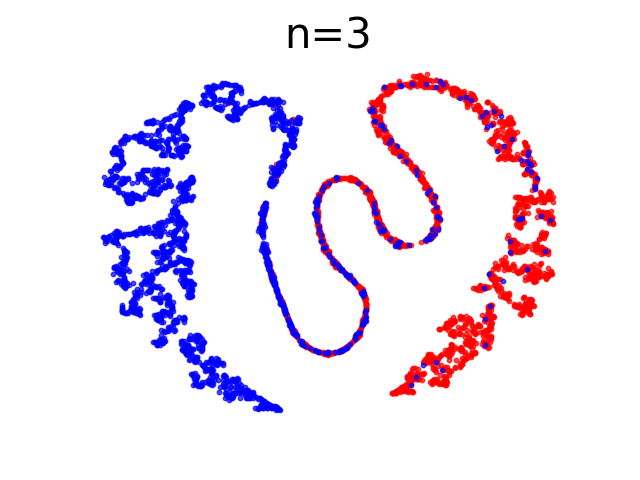}
 \end{subfigure}
  \begin{subfigure}{0.23\textwidth}
     \includegraphics[width=\textwidth]{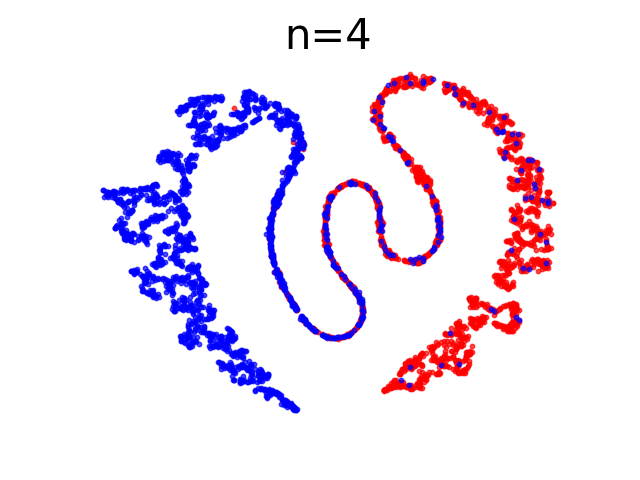}
 \end{subfigure}
  \begin{subfigure}{0.23\textwidth}
     \includegraphics[width=\textwidth]{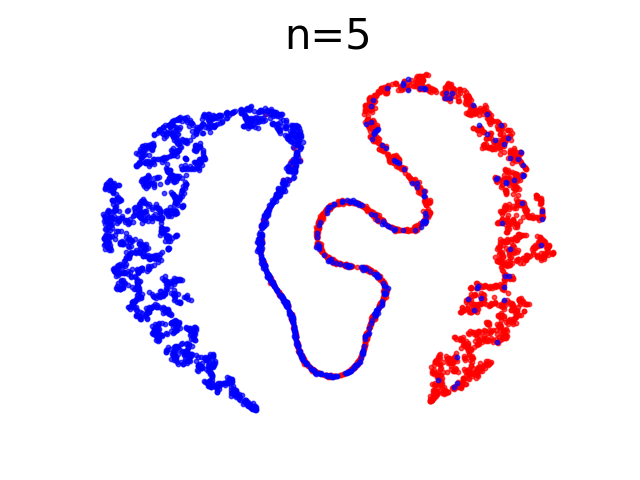}
 \end{subfigure}
  \begin{subfigure}{0.23\textwidth}
     \includegraphics[width=\textwidth]{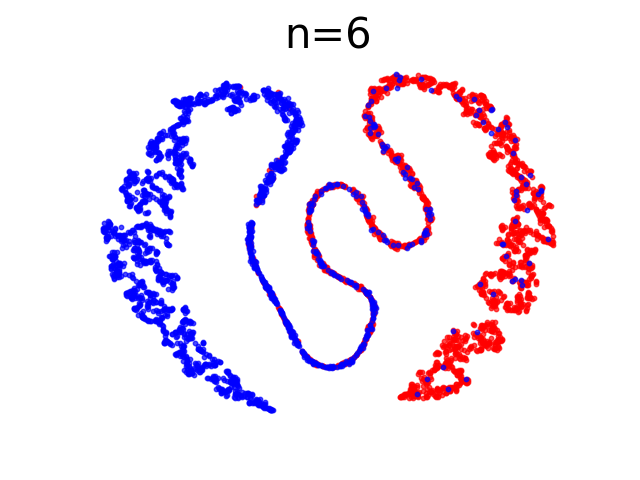}
 \end{subfigure}
  \begin{subfigure}{0.23\textwidth}
     \includegraphics[width=\textwidth]{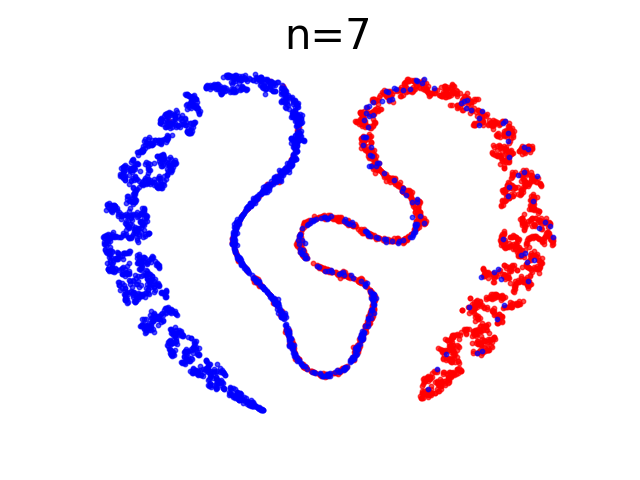}
 \end{subfigure}
  \begin{subfigure}{0.23\textwidth}
     \includegraphics[width=\textwidth]{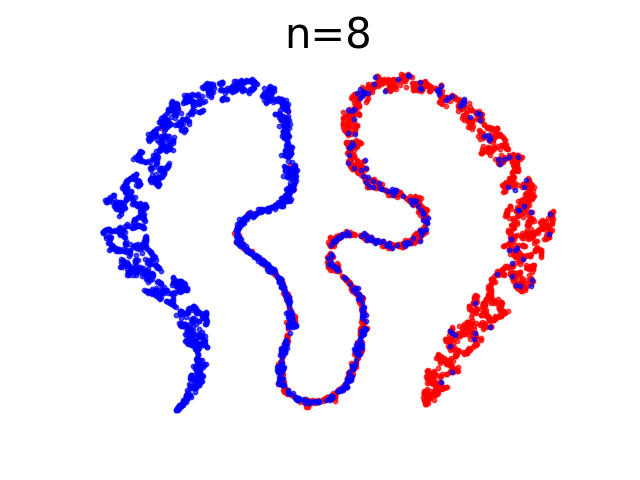}
 \end{subfigure}
  \begin{subfigure}{0.23\textwidth}
     \includegraphics[width=\textwidth]{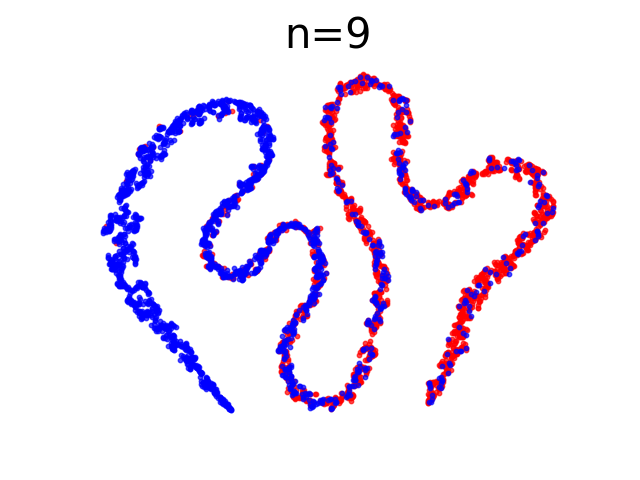}
 \end{subfigure}
  \begin{subfigure}{0.23\textwidth}
     \includegraphics[width=\textwidth]{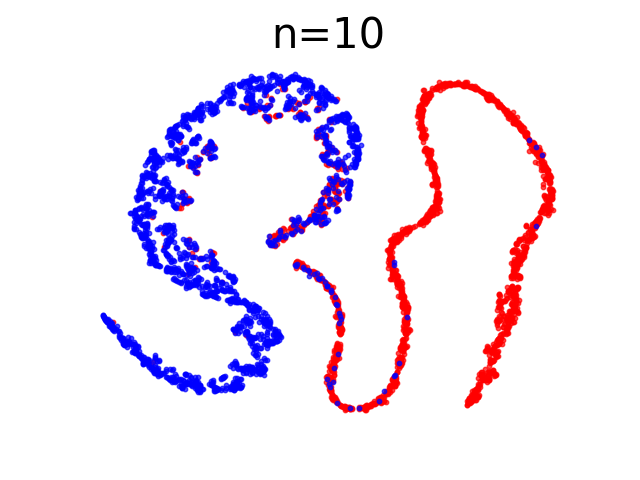}
 \end{subfigure}
 
    \caption{\textbf{TSNE of Classifier Features} 
    Red represents the forward marginals, while blue denotes the backward marginals. 
    The classifier's features can differentiate between the forward and backward marginals in most cases. However, in some instances, these marginals are indistinguishable from one another. 
    }
    \label{fig:tsne_features}
\end{figure}

\begin{figure}[H]
\centering
 \begin{subfigure}{0.23\textwidth}
     \includegraphics[width=\textwidth]{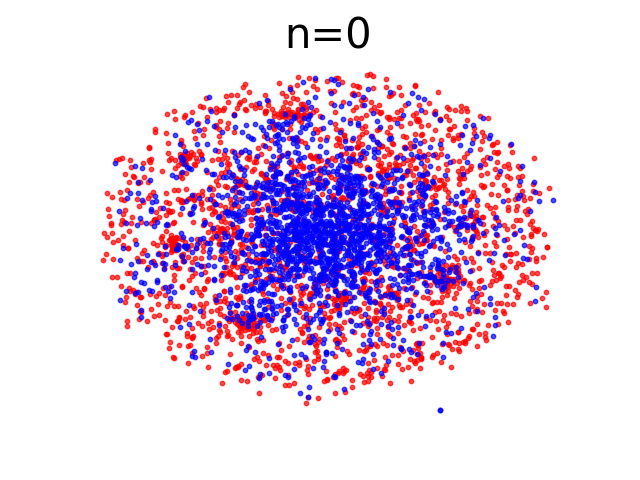}
 \end{subfigure}
 \begin{subfigure}{0.23\textwidth}
     \includegraphics[width=\textwidth]{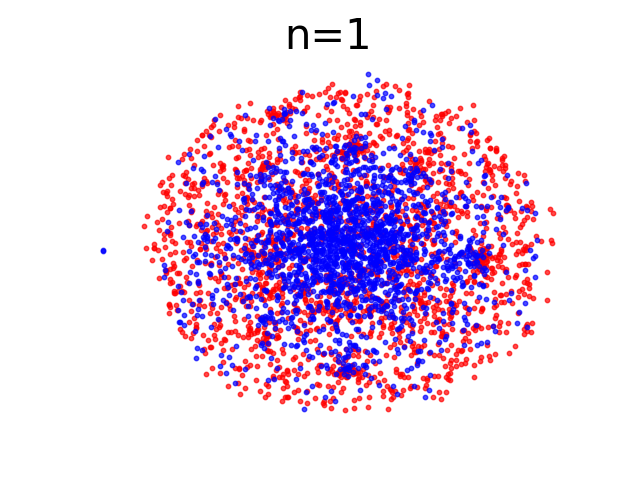}
 \end{subfigure}
 \begin{subfigure}{0.23\textwidth}
     \includegraphics[width=\textwidth]{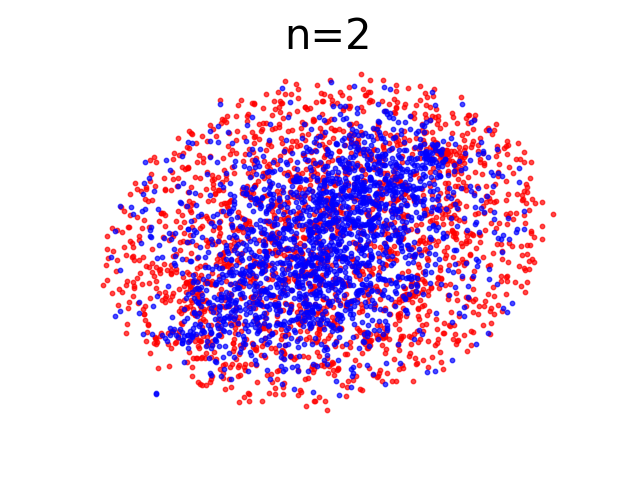}
 \end{subfigure}
 \begin{subfigure}{0.23\textwidth}
     \includegraphics[width=\textwidth]{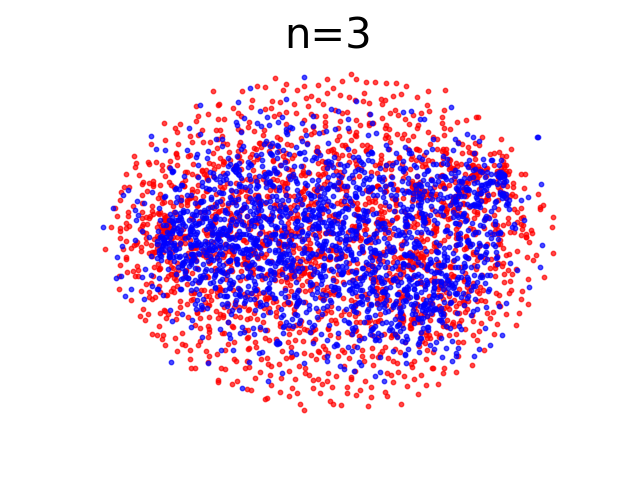}
 \end{subfigure}
  \begin{subfigure}{0.23\textwidth}
     \includegraphics[width=\textwidth]{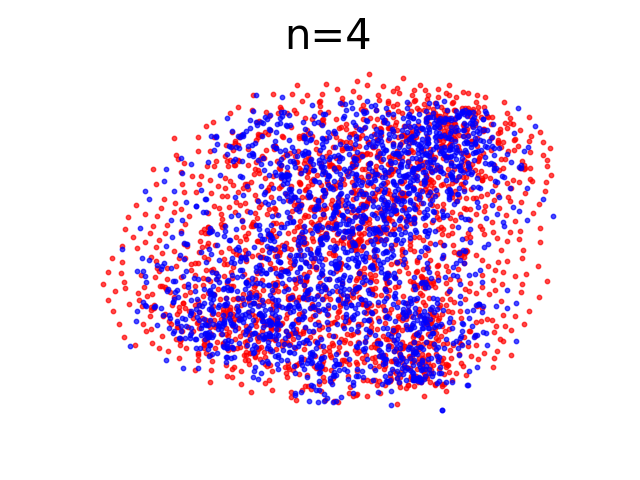}
 \end{subfigure}
  \begin{subfigure}{0.23\textwidth}
     \includegraphics[width=\textwidth]{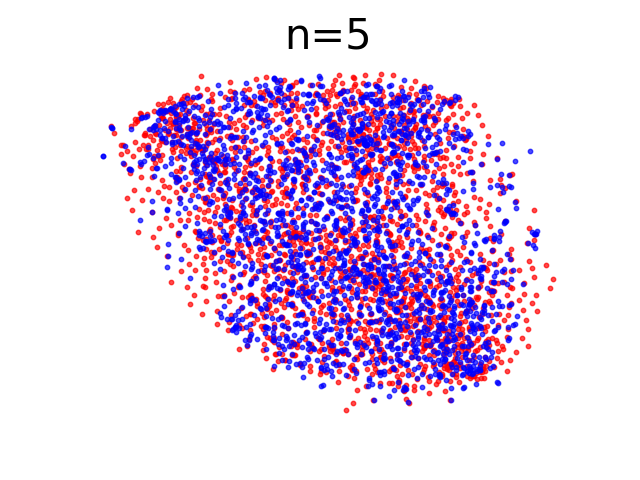}
 \end{subfigure}
  \begin{subfigure}{0.23\textwidth}
     \includegraphics[width=\textwidth]{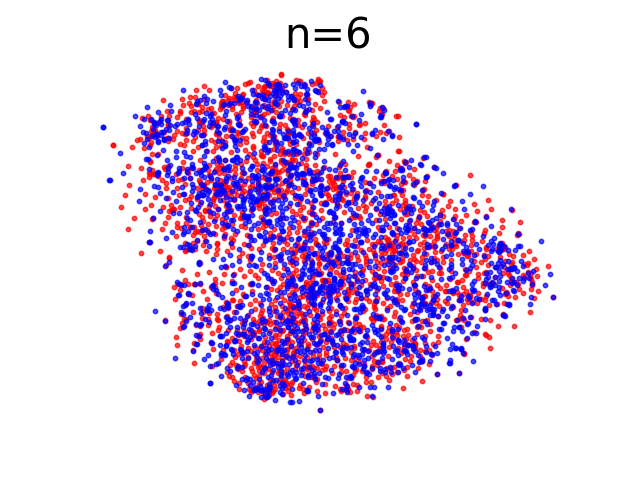}
 \end{subfigure}
  \begin{subfigure}{0.23\textwidth}
     \includegraphics[width=\textwidth]{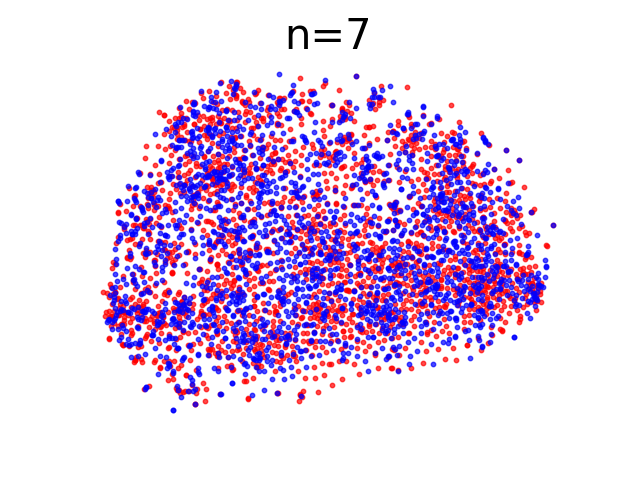}
 \end{subfigure}
  \begin{subfigure}{0.23\textwidth}
     \includegraphics[width=\textwidth]{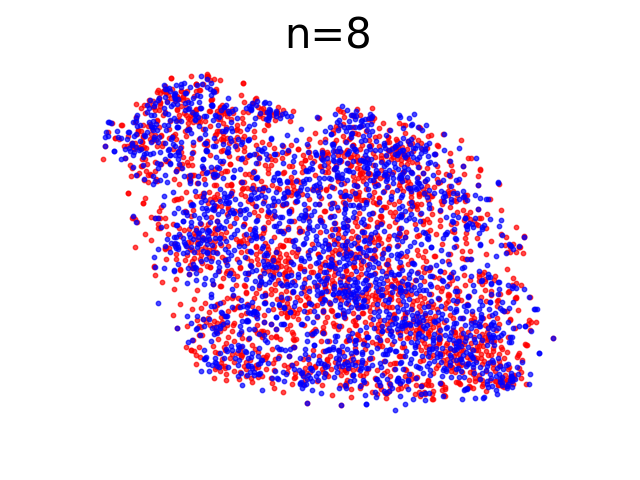}
 \end{subfigure}
  \begin{subfigure}{0.23\textwidth}
     \includegraphics[width=\textwidth]{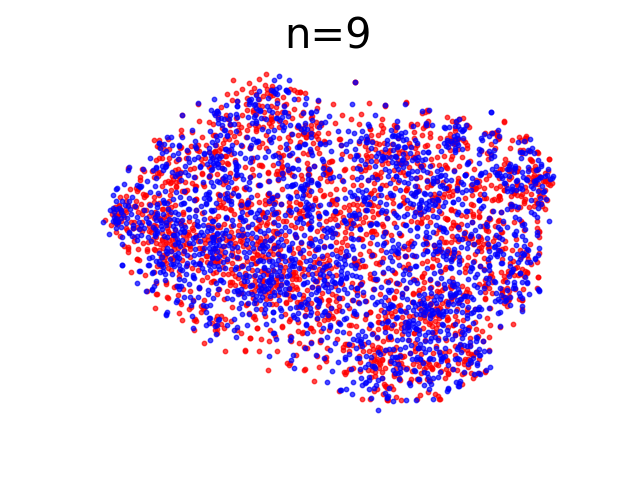}
 \end{subfigure}
  \begin{subfigure}{0.23\textwidth}
     \includegraphics[width=\textwidth]{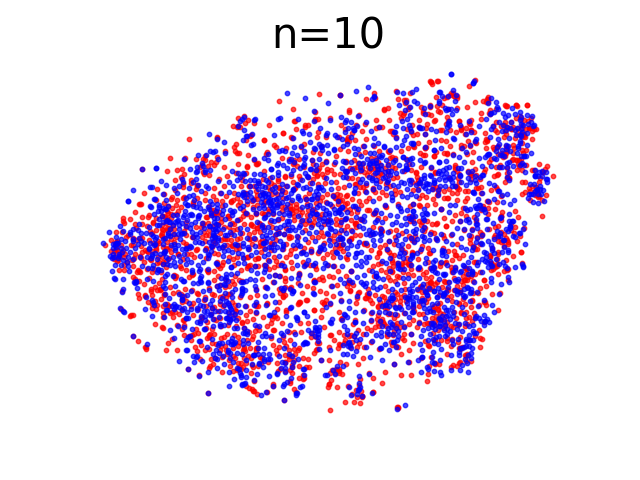}
 \end{subfigure}
 
    \caption{\textbf{TSNE of Forward and Backward Marginals}  Red represents the forward marginals, while blue denotes the backward marginals. As time progresses, we observe a convergence between these two sets of marginals. However, it's important to note that despite this increasing proximity, they never achieve perfect alignment or identity.  $n=0$ compares the approximate source distribution distribution ($p_0^b$) and Gaussian noise while $n=10$ compares the model's data approximation ($p_1^f$) and the real data.}
    \label{fig:tsne_marginals}
\end{figure}

We illustrate the distinction between the forward and backward marginals of OT-CFM flow model \citep{tong2024improving} trained on the CIFAR-10 dataset and sampled with 10-step RK4.
Fig.~\ref{fig:tsne_marginals} presents a visual representation of these marginals at each time-step $t_n$, created with t-SNE (\citet{van2008visualizing}) in order to reduce the marginals' dimensionality to two.
Fig.~\ref{fig:tsne_features} offers an alternative visualization, where the t-SNE is applied to the classifier correction model's features of the marginals at each time-step.
The visualizations demonstrate that as $n$ increases there is a trend of growing similarity between the forward and backward marginals. However, this pattern of convergence does not extend to the final time step ($n=10$), where the comparison is between the actual data and the flow model's approximation of it.

 \begin{figure}[h]%
\centering
 \includegraphics[width=0.6\textwidth]{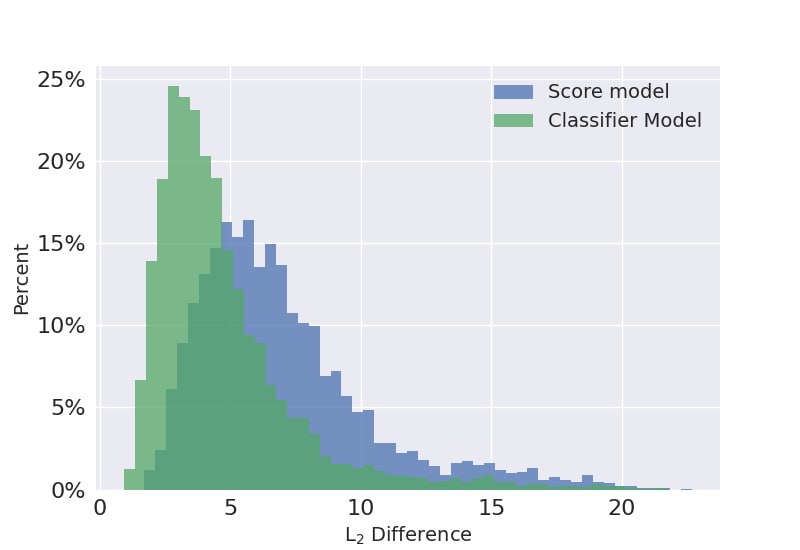}
     \caption{Percentage vs $L_2$ difference.  Histogram showing $L_2$ differences of VGG features between samples generated using a flow model alone versus with a correction model. The samples are generated from identical Gaussian noise. The percentage indicates the proportion of images with the corresponding difference. The score and classifier correction models leave most images largely unchanged, with differences concentrated in a small range, while significantly altering only outliers. This suggests that there is an intersection between the forward and backward marginals.}
\label{fig:fid_features_histogram}
\end{figure}

Reinforcing this observation, Fig.~\ref{fig:fid_features_histogram} examines the intersection of forward and backward marginals by presenting a histogram of the $L_2$ differences in VGG\footnote{VGG is the model used to calculate the FID score} features between uncorrected and corrected  samples. The distribution reveals that for both the classifier and score models, the majority of images exhibit minimal changes. This pattern indicates that corrections are primarily applied to images requiring adjustment, suggesting a close alignment between the forward and backward marginals. The selective nature of these corrections implies that the models effectively identify and address outliers, refining the overall distribution while leaving well-formed samples largely unchanged.

\section{Additional Experiments} \label{appendix:additional_experiments}

\subsection{Score Ablation Study} \label{appendix:score_ablation_study}

An ablation study on different $\sigma$ values for the score model trained on CIFAR-10 is presented in Table.~\ref{table:ablation_score}. The correction step-sizes for different sigmas were multiplied by a constant value, as the score function is multiplied by sigma during evaluation.  For all other evaluations, we employed the score model trained using $\sigma=0.005$, as it yielded superior performance compared to other values.

\begin{table}[ht]%
\centering   
\begin{tabular}{ l l l c}
 \textit{Correction Model}& NFE $\downarrow$ &C-NFE $\downarrow$ & FID $\downarrow$ \\ 
 \hline
 0.001-Score &41 &1 & 3.83 \\ 
   &43 & 3 & \textbf{3.75} \\  
   &51  & 11 & 3.77 \\  
  \hline
 0.003-Score &41 &1 & 3.51 \\ 
 &43  & 3 & \textbf{3.44} \\  
  &51  & 11 & 3.45 \\ 
     \hline
 0.005-Score & 41 &1 & 3.45 \\ 
&43 & 3 & 3.38 \\  
&51  & 11 & \textbf{3.37} \\   
    \hline
 0.01-Score & 41&1 & 3.49 \\ 
  & 43 & 3 & 3.42 \\  
  &51  & 11 & \textbf{3.38} \\ 
\hline
\end{tabular}
\caption{Ablation study on the sigma value of score correction model sampled on $n=\{[10], [0,5,10], \forall n\}$ marginals on CIFAR-10.  The $\sigma-$Score represents the value of $\sigma$ the score model was trained with.  The base flow model is OT-CFM sampled with 10-step RK4 (40 NFE). The results demonstrate that the optimal performance is achieved when $\sigma=0.005$ (the value used in the main paper).}   
\label{table:ablation_score}
\end{table}

\subsection{CIFAR-10 Classifier Guidance}\label{appendix:classifier_guidance}
\begin{figure}[ht]%
\centering
 \includegraphics[width=0.5\textwidth]{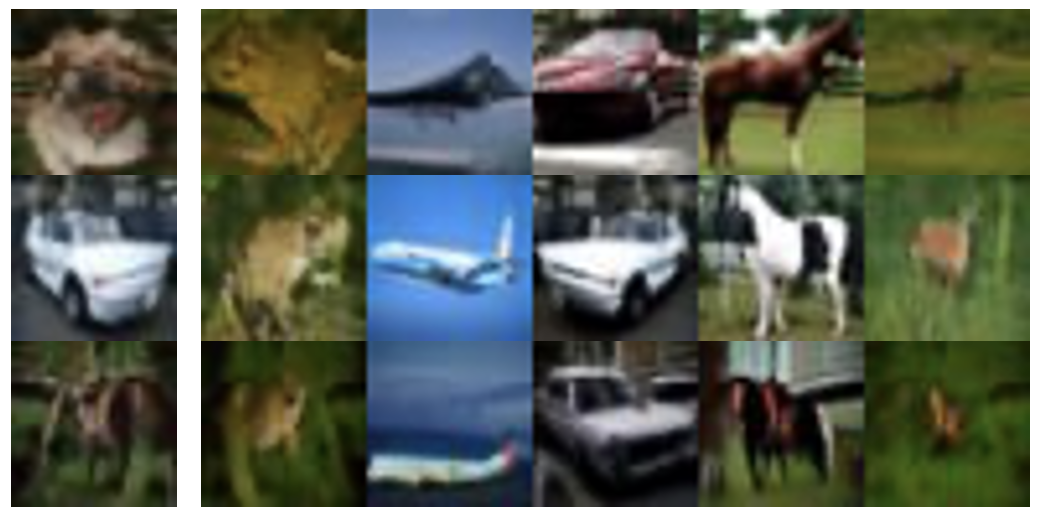}
     \caption{CIFAR-10 images generated by OT-CFM sampled with 10-step RK4, shown alongside their counterparts produced with class conditioned classifier correction.  The classifier successfully directs the noise to the correct class.}
\label{fig:class_classifier_guidance}
\end{figure}

We implement classifier guidance as described in Sec.~\ref{section:classifier} on CIFAR-10 dataset.
Fig.~\ref{fig:class_classifier_guidance} presents images produced with our corrected inference algorithm that use the class-condition classifier, where the trajectory is steered toward the correct class in the backward marginals. On the left, the original images of the flow model are presented (with no correction), and on the right the images from the same source noise, but with correction steps toward the matching classes. Even when the classifier's class matches the original class of the source noise it produces a different image of the same class, more closely aligned with the backward trajectory.

\subsection{Flow as Correction Model}
In this experiment we perform \name\ with a flow model (OT-CFM) as the correction model.  The correction is performed only on the first and last marginals with a different number of correction steps, see Table.~\ref{table:flow_correction}.  
The correction of the first marginal is done between Gaussian noise ($p_0^f$) to the approximate source distribution ($p_0^b$), and on the last marginal from the approximate target distribution ($p_1^f$) to the target distribution ($p_1^b=\pi_1$).  The first row in the table presents the FID result with no correction model, but only 10-step RK4 sampling of the OT-CFM base flow model.

The flow model only degrades the results. We hypothesize that this is due to the pairing problem, for more details refer to Appendix.~\ref{appendix:practical_considerations}.  That led us to explore other correction models - score and robust classifier.

\begin{table}[h]%
\centering   
\begin{tabular}{ l l l c}
 \textit{Correction Model}& NFE$\downarrow$ & C-NFE$\downarrow$ & FID $\downarrow$ \\ 
 \hline
 No Correction &40 & 0 & 4.34 \\
 \hline
OT-CFM -- ${1^{st}}$ \textit{Marginal} &48 &  2 & 9.74 \\  
 &64 & 6 & 6.31 \\
 &120 & 20 & 6.33  \\
  \hline
OT-CFM -- ${N^{th}}$ \textit{Marginal} &48 &  2 & 7.43 \\  
&64  &  6 & 4.39 \\
 &120 &  20 & 4.41 \\
\hline
\end{tabular}
\caption{CIFAR-10 FID comparison of OT-CFM correction and flow model trained on first and last marginals.  The correction score model was sampled with C-NFE RK4 steps, while the base flow model was sampled with 10-step RK4 (40 NFE). The correction flow model degrades the results.}
\label{table:flow_correction}
\end{table}

\subsection{CIFAR-10 Test Sample Quality}

The performance of OT-CFM flow model with and without correction on CIFAR-10 test set is presented in Table~\ref{table:test_fid}.  The correction model helps the generation quality, even though it was trained to match the training data marginals.  Where the C-NFE is greater than one, the correction steps were taken on different time-step marginals; we examine several options: for the score model the correction steps were taken on $n = {[10], [0, 5, 10], \forall n}$ and for the classifier on $n = {[8], [8, 9, 10], [6, 7, 8, 9, 10], \forall n}$, same as Table~.\ref{table:cifar10_correction_models}.

\begin{table}[h]%
\centering   
\begin{tabular}{ l l l c}
 \textit{Flow Model} & NFE $\downarrow$ & FID $\downarrow$ &\\ 
  \hline
 OT-CFM & 40 & 6.43& \\
 & 44 &6.04 & \\
 & 48 & 5.80 & \\
 & 52 & 5.70 & \\ [5mm]
 \textit{Correction Model} & NFE $\downarrow$ & C-NFE $\downarrow$ & FID $\downarrow$ \\ 

 \hline
Score &41 &  1 ~~\textit{($n=10$)} & 5.52 \\  
 &43 & 3 ~~\textit{($n=0,5,10$)} & 5.46 \\
  \hline
Classifier &42  &  1 ~~\textit{($n=8$)} & 5.66 \\  
 &46 &  3 ~~\textit{($n=8,9,10$)} & 5.57 \\
&50  &  5 ~~\textit{($n=6,7,8,9,10$)} & 5.56 \\
 \hline
\end{tabular}
\caption{CIFAR-10 test set FID performance of OT-CFM flow model with and without our correction models (score and classifier). OT-CFM is sampled without correction using 10,20, and 50-step RK4. Our correction models use OT-CFM with 10-step RK4 (40 NFE). In general, adding correction steps (C-NFE) improves the results.}
\label{table:test_fid}
\end{table}

\vspace{3ex}

\subsection{Residual Model}\label{appendix:residual_model}

Another baseline we measure against is a straightforward approach to enhance sampling, assuming we have access to the data. This involves training a residual flow matching network using the standard flow matching objective. This leverages an increase in model size effectively. The training process involves using the frozen network $v_\theta$, adding the output of the residual network ($v_\theta(t,x)+h_\psi(t,x)$), and aiming to predict the vector field (refer to Section~\ref{section:flow_models}).
Table~\ref{table:residual_net} shows the results of this approach. While it degrades the original performance, our method demonstrates improvements. We hypothesize that the residual flow matching objective is more challenging to optimize than our approach, leading to worse performance under equivalent training conditions and model sizes.

\begin{table}[h]
\centering
\begin{tabular}{ c c}

    NFE  & FID $\downarrow$   \\
 \hline
 40  & 4.44\\
 44 & 4.09\\
 48 & 3.79\\
 52 & 3.56\\
 
\hline
\end{tabular} 
\caption{Results of residual networks combined with both the base flow model and the residual network OT-CFM flow models.}
\label{table:residual_net}
\end{table}

\subsection{Flow Models RK4 Results}\label{appendix:additional_rk4}

\begin{table}[H]
\centering
\begin{tabular}{ l  c c c}

  \textit{Model} &NFE  & FID $\downarrow$ &  \\
 \cline{1-4}
 OT-CFM \citep{tong2024improving} & 40  & 4.34\\
 & 44 & 3.96\\
 & 48 & 3.73\\
 & 52 & 3.59\\
 & 56 & 3.52\\
 & 60 & 3.47\\
 I-CFM \citep{tong2024improving} & 40 & 4.29\\
 & 44 & 3.96\\
 & 48 & 3.74\\
 & 52 & 3.60\\
 & 56 & 3.52\\
 & 60 & 3.47\\
 
\hline
\end{tabular}
\caption{CIFAR-10 FID scores for OT-CFM and I-CFM flow models sampled with RK4.  The number of RK4 steps is $1/4$ of the number of the NFEs.}   
\label{table:cifar10_rk4}
\end{table}

\section{Implementation Details}\label{appendix:implementation_details} 

\subsection{Flow Models Architecture}

Our flow models architecture is based on the UNet design from \cite{nichol2021improved} that was employed in \cite{tong2024improving} and \citet{lipman2023flow}.

\begin{table}[h]
    \centering
    \begin{tabular}{lcc}
        \toprule
        & \textbf{CIFAR-10} & \textbf{ImageNet-64} \\
        \midrule
        Channels & 128 & 192 \\
        Depth & 2 & 3 \\
        Channels multiple & 1,2,2,2 & 1,2,3,4 \\
        Heads & 4 & 4 \\
        Heads Channels & 64 & 64 \\
        Attention resolution & 16 & 32,16,8 \\
        Dropout & 0.1 & 0.0 \\
        Batch size & 128 & 512 \\
        Iterations & 400K & 600k \\
        Learning Rate & 5e-4 & 1e-4 \\
        Learning Rate Scheduler & Polynomial Decay & Constant \\
        Warmup Steps & 5000 & 0 \\
        \bottomrule
    \end{tabular}
    \caption{Hyper-parameters used for CIFAR-10 and ImageNet-64 flow models.}
\end{table}

\subsection{Corrections Models Architecture}

The correction models utilized the same UNet architecture with different hyper-parameters than the flow models. The classifier model required an additional convolution layer and two linear layers to reduce the output dimension to a single scalar. The correction models were trained for $100,000$ optimization steps, however we observed convergence within $30,000$ to $40,000$ steps. The models sizes are \textbf{half} the number of trainable parameters of the original flow models, requiring \textbf{significantly} less time to converge.

\begin{table}[th]
    \centering
    \begin{tabular}{lccc}
        \toprule
        & \multicolumn{1}{c}{\textbf{Classifier}} & \multicolumn{2}{c}{\textbf{Score}} \\
        \cmidrule(r){2-2} \cmidrule(r){3-4}
        & \textbf{CIFAR-10} & \textbf{CIFAR-10} & \textbf{ImageNet-64} \\
        \midrule
        Channels & 128 & 128 & 192 \\
        Depth & 1 & 1 & 2 \\
        Channels multiple & 1,2,1,2 & 1, 2, 1, 2 & 1, 2, 3, 2 \\
        Heads & 4 & 4 & 4 \\
        Heads Channels & 32 & 32 & 64 \\
        Attention resolution & 16 & 16 & 32, 16, 8 \\
        Dropout & 0.1 & 0.1 & 0.0 \\
        Batch size & 128 & 128 & 512 \\
        Iterations & 100K & 100K & 100K \\
        Learning Rate & 0.0001 & 0.0002 & 0.0001 \\
        $\sigma$ & 0.01 & 0.005 & 0.05 \\
        Linear Layers Output & 64, 1 & - & - \\
        \bottomrule
    \end{tabular}
    \caption{Correction models hyper-parameters.}
\end{table}

\subsection{Losses}
\textbf{Score Model}: Our score model was trained using the denoising score matching loss (DSM) over the backward marginals $\{p^{b}_{t_n}\}_{n=0}^N$ with a constant noise scale $\sigma$:
\[
    L_{DSM}=\mathbb{E}\left[\mathbb{E}_{y_{t_n}\sim p_{t_n}^{b}, \epsilon \sim \mathcal{N}(0, \sigma^2 I)}
        \left[\left\|s_{\psi}(t_n,y_{t_n} + \epsilon) + \epsilon \right\|_2^2\right]\right]
\]
where $s_{\psi}(t_n,y_{t_n})$ is the score model parameterized by $\psi$.
For more details see Sec.~\ref{section:score_models}.

\textbf{Classifier Model}: Our robust classifier was trained using $L_{\text{classifier}}$ which is comprised of 3 terms:
\[
        L_{\text{classifier}} = L_{BCE} + L_{AT} + L_{GA}
\]

\textbf{Binary Cross-Entropy}: The binary cross-entropy (BCE) is used to distinguish between the backward marginals $p^b_{t_n}$ and the forward marginals $p^f_{t_n}$:
\[
        L_{BCE} = \mathbb{E}\left[\mathbb{E}_{y_{t_n} \sim p_{t_n}^b, x_{t_n} \sim p_{t_n}^f}\left[BCE(c_{\psi}(t_n,y_{t_n}), 1) + BCE(c_{\psi}(t_n,x_{t_n}), 0)\right]\right]
\]
where $c_{\psi}(t_n,y)$ is the classifier output for input $y$ at time $t_n$, parameterized by $\psi$.\\
\textbf{Adversarial Training (AT)}: 
To enhance its robustness, the classifier was also trained on adversarial examples. This process involves:
\begin{enumerate}
    \item Sampling $y_{t_n} \sim p^b_{t_n}$ and $x_{t_n} \sim p^f_{t_n}$.
    \item Optimizing norm-clipped additive perturbations $\eta_y$ and $\eta_x$ to:
        \begin{itemize}
            \item Decrease the classifier value for $y$: $c_{\psi}(t_n, y_{t_n} + \eta_y)$
            \item Increase the classifier value for $x$: $c_{\psi}(t_n, x_{t_n} + \eta_x)$
        \end{itemize}
\end{enumerate}
For a more detailed explanation see Sec.~\ref{appendix:classifier_training}.
\[
        L_{AT} = \mathbb{E}\left[\mathbb{E}_{y_{t_n} \sim p_{t_n}^b, x_{t_n} \sim p_{t_n}^f}\left[BCE(c_{\psi}(t_n,y_{t_n} + \eta_y), 1) + BCE(c_{\psi}(t_n,x_{t_n} + \eta_x), 0)\right]\right]
\]
\textbf{Gradient Alignment (GA)}:
To align the classifier's gradient near the backward marginals, we introduce a small amount of Gaussian noise and use cosine similarity to adjust the classifier's gradient in the direction opposite to the noise (towards the backward marginal).
\[
L_{GA} = \mathbb{E}\left[\mathbb{E}_{y_{t_n} \sim p_{t_n}^b, \epsilon \sim \mathcal{N}(0, \sigma^2 I)}\left[1 - \frac{\langle \nabla_{y_{t_n}}c_{\psi}(t_n, x_{t_n} + \epsilon), \epsilon \rangle}{\|\nabla_{y_{t_n}}c_{\psi}(t_n, x_{t_n} + \epsilon)\|_2 \|\epsilon\|_2}\right]\right]
\]

\subsection{Evaluation Parameters}

\noindent\textbf{Hyper-parameters:}

\textit{Step size $\alpha$:} The configuration with the best FID was selected from a grid search over the interval $[0,2]$ with a step size of 0.05. For the classifier's final step, the grid search was conducted over the interval $[0,0.1]$ with a step size of 0.01.

\textit{Noise $\beta$:} The configuration with the best FID was selected from a grid search over the interval $[0,0.1]$ with a step size of 0.01.

\noindent\textbf{Sampling:}

The hyper-parameters for all evaluations are described below, except for parallel sampling, where the time-steps are shifted by 1 ($10 \rightarrow 9$, $9\rightarrow 8$, etc).  Additionally, for parallel classifier the final step-size is $0.02$ instead of $0.06$.
 
\begin{table}[htbp!]
    \centering
    \begin{tabular}{>{\centering\arraybackslash}p{1.0cm}|p{1.3cm}@{}p{0.75cm}|p{1.3cm}@{}p{0.75cm}|p{1.3cm}@{}p{0.75cm}|p{1.3cm}@{}p{0.75cm}}
        \toprule
        \textbf{Corr. Step} & \multicolumn{8}{c}{\textbf{Classifier}} \\
        \cmidrule(lr){2-9}
        & \multicolumn{2}{c|}{\textbf{10 Steps}} & \multicolumn{2}{c|}{\textbf{5 Steps}} & \multicolumn{2}{c|}{\textbf{3 Steps}} & \multicolumn{2}{c}{\textbf{1 Step}} \\
        \cmidrule(lr){2-9}
        & \textbf{Step Size} & \textbf{Noise} & \textbf{Step Size} & \textbf{Noise} & \textbf{Step Size} & \textbf{Noise} & \textbf{Step Size} & \textbf{Noise} \\
        \midrule
        0  & 1.5 & 0.05 & - & - & - & - & - & - \\
        1  & 1.5 & 0.0  & - & - & - & - & - & - \\
        2  & 1.5 & 0.0  & - & - & - & - & - & - \\
        3  & 1.5 & 0.0  & - & - & - & - & - & - \\
        4  & 1.5 & 0.0  & - & - & - & - & - & - \\
        5  & 1.5 & 0.0  & - & - & - & - & - & - \\
        6  & 1.5 & 0.08 & 1.5 & 0.08 & - & - & - & - \\
        7  & 1.5 & 0.0  & 1.5 & 0.0  & - & - & - & - \\
        8  & 1.5 & 0.0  & 1.5 & 0.0  & 1.0 & 0.05 & 1.0 & 0.05 \\
        9  & 1.5 & 0.0  & 1.5 & 0.0  & 0.4 & 0.0  & - & - \\
        10 & 0.06 & 0.0  & 0.06 & 0.0  & 0.06 & 0.0  & - & - \\
        \bottomrule
    \end{tabular}
    \caption{CIFAR-10 evaluation hyper-parameters for the classifier correction model.}
    \label{tab:classifier-parameters}
\end{table}

\begin{table}[htbp!]
    \centering
    \begin{tabular}{>{\centering\arraybackslash}m{1.2cm}|cc|cc|cc}
        \toprule
        \textbf{Corr. Step} & \multicolumn{6}{c}{\textbf{Score}} \\
        \cmidrule(lr){2-7}
        & \multicolumn{2}{c|}{\textbf{10 Steps}} & \multicolumn{2}{c|}{\textbf{3 Steps}} & \multicolumn{2}{c}{\textbf{1 Step}} \\
        \cmidrule(lr){2-7}
        & \textbf{Step Size} & \textbf{Noise} & \textbf{Step Size} & \textbf{Noise} & \textbf{Step Size} & \textbf{Noise} \\
        \midrule
        0  & 0.4  & 0.03 & 0.35 & 0.03 & - & - \\
        1  & 0.3  & 0.0  & -    & -    & - & - \\
        2  & 0.3  & 0.0  & -    & -    & - & - \\
        3  & 0.3  & 0.0  & -    & -    & - & - \\
        4  & 0.3  & 0.01 & -    & -    & - & - \\
        5  & 0.35 & 0.05 & 0.45 & 0.05 & - & - \\
        6  & 0.3  & 0.0  & -    & -    & - & - \\
        7  & 0.3  & 0.0  & -    & -    & - & - \\
        8  & 0.3  & 0.01 & -    & -    & - & - \\
        9  & 0.3  & 0.00 & -    & -    & - & - \\
        10 & 2.0  & 0.0  & 2.0  & 0.0  & 2.0 & 0.0 \\
        \bottomrule
    \end{tabular}
    \caption{CIFAR-10 evaluation hyper-parameters for score correction model.}
    \label{tab:score-parameters}
\end{table}

\begin{table}[htbp!]
    \centering
    \begin{tabular}{>{\centering\arraybackslash}m{1.2cm}|cc|cc|cc}
        \toprule
        \textbf{Corr. Step} & \multicolumn{6}{c}{\textbf{Score}} \\
        \cmidrule(lr){2-7}
        & \multicolumn{2}{c|}{\textbf{5 Steps}} & \multicolumn{2}{c|}{\textbf{2 Steps}} & \multicolumn{2}{c}{\textbf{1 Step}} \\
        \cmidrule(lr){2-7}
        & \textbf{Step Size} & \textbf{Noise} & \textbf{Step Size} & \textbf{Noise} & \textbf{Step Size} & \textbf{Noise} \\
        \midrule
        0  & 0.4 & 0.1 & 0.4 & 0.01 & - & - \\
        1  & 0.2 & 0.05 & -   & -    & - & - \\
        2  & 0.2 & 0.0 & -   & -    & - & - \\
        3  & 0.2 & 0.0 & -   & -    & - & - \\
        4  & 0.2 & 0.05 & -   & -    & - & - \\
        5  & 0.4 & 0.0 & 0.4 & 0.0  & 0.4 & 0.0 \\
        \bottomrule
    \end{tabular}
    \caption{ImageNet-64 evaluation hyper-parameters for score correction model.}
    \label{tab:score-parameters-imagenet}
\end{table}

The score model's Langevin Dynamics sampling used $50$ and $100$ steps with step sizes of $0.1$ and $0.05$, respectively, yielding the best results among step sizes $[0.01, 0.05, 0.1]$.

\vspace{20mm}

\section{Qualitaive Examples} \label{appendix:imagenet_images}
\vspace{-10pt}

\begin{figure}[htbp!]

\centering
  \begin{subfigure}{0.47\textwidth}
     \includegraphics[width=\textwidth]{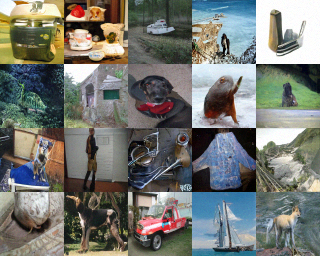}
     \caption{OT-CFM}
    \end{subfigure}

    \begin{subfigure}{0.47\textwidth}
     \includegraphics[width=\textwidth]{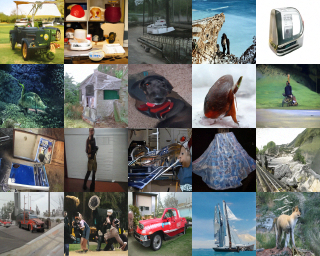}
      \caption{Score Correction Model}
    \end{subfigure}

    \begin{subfigure}{0.47\textwidth}
     \includegraphics[width=\textwidth]{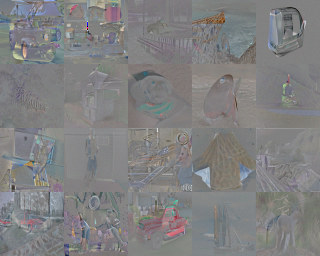}
     \caption{Difference}
    \end{subfigure}
 
    \caption{ImageNet-64 image generation using OT-CFM alone and with score correction model. The corrected images demonstrate enhanced sharpness and definition compared to their uncorrected counterparts.}
    \label{fig:imagenet_comparisons}
\end{figure}

\begin{figure}[htbp!]
 \label{appendix:cifar_images}
\centering
  \begin{subfigure}{0.47\textwidth}
     \includegraphics[width=\textwidth]{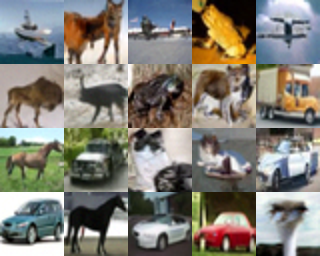}
     \caption{OT-CFM}
    \end{subfigure}

    \begin{subfigure}{0.47\textwidth}
     \includegraphics[width=\textwidth]{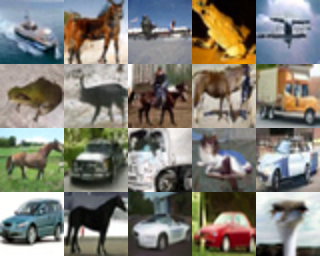}
      \caption{Score Correction Model}
    \end{subfigure}

    \begin{subfigure}{0.47\textwidth}
     \includegraphics[width=\textwidth]{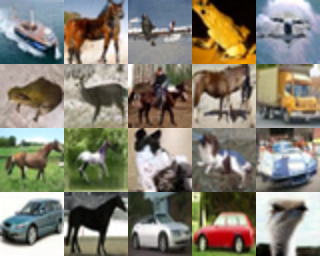}
     \caption{Classifier Correction Model}
    \end{subfigure}
 
    \caption{CIFAR-10 image generation using OT-CFM alone and with score correction and classifier correction models. Despite the distinct nature of these correction models, their suggested improvements often exhibit remarkable similarity. The corrected images demonstrate enhanced sharpness and definition compared to their uncorrected counterparts.}
    \label{fig:cifar_comparisons}
\end{figure}

\begin{figure}[htbp!]

\centering
  \begin{subfigure}{0.45\textwidth}
     \includegraphics[width=\textwidth]{new_images/imagenet/flow_imagenent.png}
     \caption{OT-CFM}
    \end{subfigure}

    \begin{subfigure}{0.45\textwidth}
     \includegraphics[width=\textwidth]{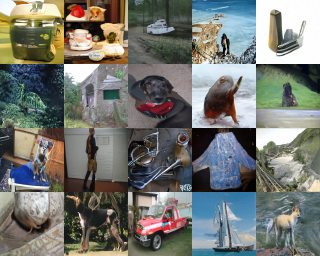}
        \caption{Last Marginal Score Correction Model}
    \end{subfigure}

    \begin{subfigure}{0.45\textwidth}
     \includegraphics[width=\textwidth]{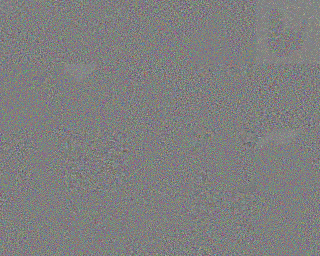}
        \caption{Correction Difference}
    \end{subfigure}
 
    \caption{ImageNet-64 images generated by OT-CFM alone and with score correction model applied solely to the final step, accompanied by their difference (amplified for visibility). The noise-like appearance of the difference suggests the presence of residual noise in the model's predictions.}
    \label{fig:imagenet_last_step_score_comparison}
\end{figure}

\begin{figure}[htbp]
\centering

  \begin{subfigure}{0.45\textwidth}
     \includegraphics[width=\textwidth]{new_images/cifar/cifar_regular_images.png}
     \caption{OT-CFM}
    \end{subfigure}

    \begin{subfigure}{0.45\textwidth}
     \includegraphics[width=\textwidth]{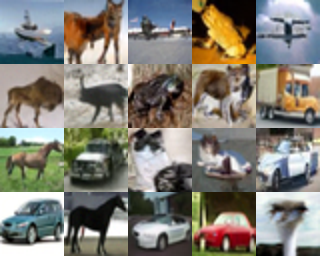}
        \caption{Last Marginal Score Correction Model}
    \end{subfigure}

    \begin{subfigure}{0.45\textwidth}
     \includegraphics[width=\textwidth]{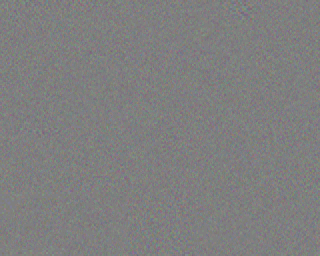}
        \caption{Correction Difference}
    \end{subfigure}
 
    \caption{CIFAR-10 images generated by OT-CFM alone and with score correction model applied solely to the final step, accompanied by their difference (amplified for visibility). The noise-like appearance of the difference suggests the presence of residual noise in the model's predictions.}
    \label{fig:cifar_last_step_score_comparison}
\end{figure}

\end{document}